\documentclass[10pt]{article} 
\usepackage[preprint]{tmlr}

\usepackage{amsmath,amsfonts,bm}

\def\eqref#1{equation~\ref{#1}}

\def\1{\bm{1}}

\def\rmA{{\mathbf{A}}}

\def\rmD{{\mathbf{D}}}
\def\rmE{{\mathbf{E}}}

\def\rmH{{\mathbf{H}}}

\def\rmM{{\mathbf{M}}}

\def\rmV{{\mathbf{V}}}
\def\rmW{{\mathbf{W}}}
\def\rmX{{\mathbf{X}}}

\def\ve{{\bm{e}}}

\def\vh{{\bm{h}}}

\def\vm{{\bm{m}}}

\def\vs{{\bm{s}}}

\def\vw{{\bm{w}}}
\def\vx{{\bm{x}}}

\DeclareMathAlphabet{\mathsfit}{\encodingdefault}{\sfdefault}{m}{sl}
\SetMathAlphabet{\mathsfit}{bold}{\encodingdefault}{\sfdefault}{bx}{n}

\newcommand{\R}{\mathbb{R}}

\usepackage{hyperref}
\usepackage{url}

\usepackage{booktabs}						
\usepackage{multirow}						
\usepackage{amsfonts}						
\usepackage{graphicx}						

\usepackage{comment}
\usepackage{subcaption}
\usepackage{amsmath}
\usepackage{wrapfig}
\usepackage[ruled,vlined]{algorithm2e}
\usepackage[normalem]{ulem}

\definecolor{royalblue}{rgb}{0.2549, 0.4118, 0.8824}
\definecolor{slide}{rgb}{0.8118, 0.4157, 0.5961}
\definecolor{region}{rgb}{0.631, 0.647, 0.847}
\definecolor{cell}{rgb}{0.345, 0.145, 0.353}

\hypersetup{
    colorlinks=true,  %
    citecolor=royalblue, 
    linkcolor=royalblue,
    filecolor=royalblue,
    urlcolor=royalblue,
}

\title{GrapHist: Graph Self-Supervision for Histopathology}

\author{\name Sevda Öğüt\thanks{Co-first authorship.} \email \href{mailto:sevda.ogut@epfl.ch}{sevda.ogut@epfl.ch} \\
      \addr EPFL, Switzerland
      \AND
      \name Cédric Vincent-Cuaz\footnotemark[1] \email cedric.vincent-cuaz@unibe.ch \\
      \addr Department for Biomedical Research, University of Bern, Switzerland \\
      \addr Swiss Institute of Bioinformatics, Switzerland
      \AND
      \name Tomas Gadea Alcaide\thanks{Co-second authorship. Work done while at EPFL.} \email tomas.gadeaalcaide@uzh.ch \\
      \addr University of Zurich, Switzerland   
      \AND
      \name Mahbod Tajdini\footnotemark[2] \email mahbod.tajdini@uzh.ch \\
      \addr University of Zurich, Switzerland      
      \AND
      \name Carlos Hurtado Comin\footnotemark[2] \email carloshurtadocomin@gmail.com \\
      \addr Mistral AI, France
      \AND
      \name Natalia Dubljevic\footnotemark[2] \email natalia.dubljevic@gmail.com \\
      \addr Canada
      \AND
      \name Vaishnavi Subramanian\footnotemark[2] \email svaishnavi411@gmail.com \\
      \addr India
      \AND
      \name Pascal Frossard \email pascal.frossard@epfl.ch \\
      \addr EPFL, Switzerland
      \AND
      \name Dorina Thanou \email dorina.thanou@epfl.ch \\
      \addr EPFL, Switzerland}

\def\month{MM}  
\def\year{YYYY} 
\def\openreview{\url{https://openreview.net/forum?id=XXXX}} 

\begin{document}

\maketitle

\begin{abstract}
Self-supervised vision models have achieved strong performance in digital pathology, but their representations are learned from regular image grids without explicitly encoding the cellular organization that structures histopathological tissue.
We investigate whether representing tissues directly as cell graphs can provide a more data- and compute-efficient inductive bias for histopathology representation learning.
We introduce \textbf{GrapHist}, a graph-based self-supervised learning framework that combines masked graph autoencoding with adaptive message passing to learn transferable cell-centric representations.
To examine how much biological context to retain in this representation, we consider two configurations: \textbf{GrapHist-S}, which models cellular morphology and spatial organization, and \textbf{GrapHist-E}, which additionally incorporates descriptors of tissue context between neighboring cells.
We pre-train both variants on a large collection of 11 million cell graphs derived from breast tissues and evaluate across cell-, region-, and slide-level benchmarks, including external-cohort and cross-tissue transfer.
Under matched source data and training exposure, GrapHist-S consistently outperforms DINOv2- and MAE-based vision baselines across the evaluated downstream tasks while using substantially fewer parameters and less pre-training time.
GrapHist-E provides additional, task-dependent gains, indicating that intercellular context can complement cellular morphology and topology in selected settings.
Finally, we release the processed graph representations of five public digital pathology datasets, together with our \href{https://github.com/ogutsevda/graphist/}{code}, to facilitate reproducible evaluation of graph methods in computational pathology.
\end{abstract}
\section{Introduction}

Recent advances in large-scale self-supervised learning (SSL) have led to promising vision foundation models for digital pathology, addressing clinically relevant tasks such as cancer typing and grading, treatment response assessment, and survival prediction~\citep{wang2022transformer, Filiot2023ScalingSSLforHistoWithMIM, chen2024towards}.
Most contemporary approaches operate directly on image patches extracted from whole-slide images (WSIs), typically using vision transformers to represent each patch as a sequence of regularly spaced visual tokens~\citep{dosovitskiy2020image}.
Scaling these models across increasingly large and diverse pathology cohorts has yielded substantial gains, but also requires considerable amounts of data and computation.
Moreover, these architectures use tokens defined by the image grid rather than the cellular organization that constitutes a fundamental level of tissue structure in histopathology~\citep{shafi2023artificial,chen2022scaling}.

Cell graphs provide a complementary representation in which tissue organization is encoded explicitly.
Individual cells can be represented as nodes characterized by morphological, intensity, and texture descriptors, while edges encode their spatial or relational organization.
Such representations have been successfully used for task-specific prediction in computational pathology, including tumor grading, treatment-response prediction, and region classification~\citep{zhou2019cgc,pati2022hierarchical,yang2024prediction}.
By abstracting raw images into cellular constituents and their relationships, cell graphs introduce a strong structural prior and substantially reduce the number of elements the downstream model must process.
This contrasts with vision models, which retain dense image information but leave the identification of biologically meaningful cellular structure to the learning process.
Conversely, the cell-graph abstraction is more selective: common constructions primarily emphasize cellular attributes and spatial organization, while image information associated with the tissue between neighboring cells is not necessarily retained.
Whether this additional local context provides discriminative information beyond cellular morphology and topology therefore remains an important question.
Despite the increasing use of graph neural networks (GNNs) in computational pathology, existing approaches are predominantly trained for individual downstream tasks and on comparatively limited datasets.
In contrast, the potential of large-scale self-supervised pre-training on cell graphs remains largely unexplored.

To address this, we propose \textbf{GrapHist}, a large-scale self-supervised pre-training framework that operates directly on histopathology cell graphs and learns transferable cell-centric representations across biological scales.
To examine how much tissue context should be retained by this abstraction, we study two configurations. \textbf{GrapHist-S} represents cells through discriminative shape, color-intensity, and texture features, with edges encoding their spatial organization. \textbf{GrapHist-E} extends this representation by augmenting spatial relationships with descriptors of the intercellular image context, thereby recovering information discarded by the purely cell-centric abstraction.
We stress that GrapHist-S is the parsimonious configuration, whereas GrapHist-E trades additional computation for richer tissue context and improved performance on context-sensitive tasks.
Both versions learn generalizable embeddings through self-supervised pre-training using masked autoencoding of cell graphs, following the GraphMAE framework~\citep{hou2022graphmae}.
In this setting, subsets of node features are masked and then reconstructed from their local neighborhoods.
Cellular neighborhoods in histopathology can exhibit markedly different local organization: homogeneous regions coexist with heterogeneous interfaces between distinct cell populations, particularly within the tumor microenvironment~\citep{anderson2020tumor}.
We therefore adopt Adaptive Channel Mixing (ACM)~\citep{luan2022revisiting}, which allows each cell to adaptively combine low-pass, high-pass, and identity signals rather than imposing a single smoothing regime across the graph.

We pre-train both versions on approximately 11 million cell graphs derived from breast tissue images and evaluate their representations across cell phenotyping, region- and slide-level tumor subtyping, and patient survival analysis, including external-cohort and cross-tissue transfer.
We pre-train DINOv2- and MAE-based vision baselines from scratch using the same source images and matched training exposure~\citep{oquab2023dinov2, he2022masked}.
GrapHist-S outperforms these baselines across the evaluated downstream settings while using a substantially smaller encoder and lower model-side pre-training cost.
GrapHist-E provides additional, task-dependent gains, showing that richer intercellular context can be beneficial in selected settings.
Comparisons with fully supervised graph baselines further assess the contribution of self-supervised pre-training.
Finally, we publicly release five graph-enhanced digital pathology datasets to support reproducible research.

Our main contributions are as follows:

\begin{itemize}
    \item We introduce \textbf{GrapHist}, a large-scale self-supervised framework for learning transferable histopathology representations directly from cell graphs, combining masked graph autoencoding with adaptive message passing and supporting different levels of cellular and intercellular context.
    \item Under matched source data and training exposure, we show that the parsimonious GrapHist-S configuration consistently outperforms the evaluated DINOv2- and MAE-based vision baselines across cell-, region-, and slide-level tasks, while GrapHist-E reveals task-dependent benefits from retaining richer intercellular tissue context.
    \item Comparisons with fully supervised graph models reveal how the value of pre-training changes with source--target shift: GrapHist variants remain competitive with or outperform matched supervised models in breast-aligned settings, whereas models combining target supervision with broad pathology foundation-model features become comparatively stronger under cross-tissue transfer.
    Parameter-efficient adaptation substantially reduces this gap.
\end{itemize}

\section{Related work}

\subsection{Foundation models for digital pathology}

Computational pathology has shifted from natural-image transfer toward self-supervised pre-training on large histopathology collections, a rapidly expanding landscape summarized in recent surveys~\citep{xiong2025survey,li2026survey}.
A dominant line of work combines large Vision Transformers with DINO-, iBOT-, or DINOv2-like objectives and increasingly large and diverse datasets, now reaching millions of WSIs and billion-parameter models~\citep{wang2022transformer, xu2024whole,vorontsov2024foundation,chen2024towards, nechaev2024hibou,zimmermann2024virchow2}.
In parallel, recent work has shown that performance also depends strongly on the training regime: Midnight improves data efficiency through a pathology-adapted DINOv2 recipe and high-resolution post-training~\citep{karasikov2025training}, while other approaches optimize corpus composition through automatic curation and balanced sampling~\citep{chen2025revisiting}.
Together, these developments show that pathology foundation models benefit from both scale and better use of available data.
Despite these advances, image-based models largely retain the same representational principle: tissue is encoded from regular pixel grids, leaving cellular identity and spatial organization to emerge implicitly during pre-training.
This paradigm is highly effective, with leading models showing strong but task-dependent performance~\citep{neidlinger2025benchmarking}.
Our work investigates an orthogonal axis: whether explicitly representing cells and their spatial relationships as graphs provides a useful structural inductive bias for self-supervised learning.
Scaling, curation, sampling, and augmentation remain compatible with this formulation.
Here, we deliberately isolate the effect of the cellular graph representation itself.

\subsection{Graphs in histopathology}

A growing body of work has represented histopathology images as graphs to explicitly model spatial tissue organization at different biological scales.
GRAPE-Net~\citep{gindra2024graph} represents image patches as graph nodes for lung tumor stratification, whereas methods such as OCDPI~\citep{yang2024prediction} and HACT-Net~\citep{pati2022hierarchical} construct cellular or hierarchical cell-to-tissue graphs for treatment-response prediction and tumor classification.
More closely related to our setting,~\citet{pina2022self} learn self-supervised representations from local cell graphs and subsequently organize them into higher-level WSI graphs for region-of-interest discovery.
Recent work has further explored self-supervised graph pre-training in pathology using tissue-entity graphs, patch-level WSI graphs, or hypergraphs, including H-MGDM, PGSL, and DMCL-HFM~\citep{zhuang2025dynamic, li2025pathological, zhou2026dual}.
These methods establish the relevance of graph self-supervision for histopathology, but primarily learn representations at the tissue-region or WSI level, often within task-specific or slide-level pipelines.
In contrast, GrapHist focuses on large-scale pre-training of a single \emph{cell-graph} encoder whose node representations can be transferred directly across cell-, region-, and slide-level tasks.
We refer to~\citet{brussee2025graph} for a broader survey of graph learning in histopathology.

\subsection{Graph self-supervised learning}

Graph self-supervised learning is commonly approached through contrastive or reconstruction-based objectives~\citep{liu2022graph}.
Contrastive methods such as GRACE~\citep{zhu2020deep} and GraphCL~\citep{you2020graph} enforce consistency across perturbed graph views, whereas GraphMAE~\citep{hou2022graphmae} reconstructs masked node attributes from their context.
We adopt the latter because cell-graph topology has explicit spatial meaning, making structural augmentations difficult to define without additional biological assumptions.
While extensions such as GraphMAE2~\citep{hou2023graphmae2} introduce multi-view re-masking and latent prediction, we retain the simpler reconstruction objective to isolate the effects of graph representation and message passing, while maintaining a conceptually closer comparison with masked-autoencoding vision baselines.

Recent work has begun to explore graph SSL under heterophily through similarity-based graph decompositions, frequency-biased views, or specialized contrastive objectives~\citep{yang2024graph,wang2024hetergcl,wan2024s3gcl}.
Such approaches encode assumptions about which relations should be smoothed or contrasted.
Notably, HLCL infers these regimes directly from similarities between input node features.
This is less attractive for our setting, where initial morphological descriptors need not be well separated according to biological identity in their Euclidean feature space.
GrapHist instead keeps the SSL objective generic and uses Adaptive Channel Mixing~\citep{luan2022revisiting} to learn node-dependent mixtures of low-pass, high-pass, and identity signals over the same spatial graph, accommodating homogeneous and heterogeneous cellular regions without pre-assigning either regime.
\begin{figure}[t]
    \centering
    \includegraphics[width=0.80\linewidth]{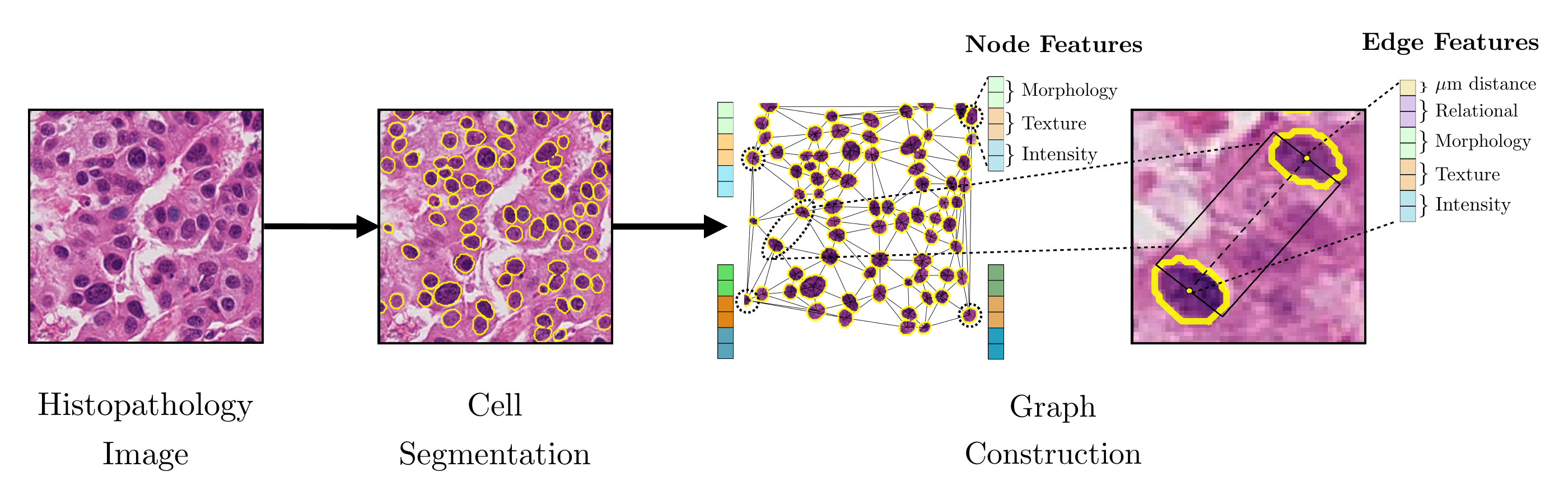}
    \caption{\textbf{Pre-processing steps.} Individual cells are segmented within a digital pathology image. These cells and their spatial arrangement are then converted into a graph. Each cell is a node with morphology, texture, and intensity descriptors. Edges connect neighboring cells and are weighted by their geographic distance, and optionally with additional features. Best viewed in color.}
    \label{fig:preprocessing}
\end{figure}

\section{GrapHist: self-supervised learning for digital pathology with cell graphs}
\label{sec:methodology}

GrapHist learns transferable histopathology representations by performing self-supervised pre-training directly on cell graphs.
The framework consists of three stages.
First, histopathology images are converted into attributed cell graphs that encode cellular morphology and spatial organization, and optionally, intercellular context (Section~\ref{sec:preprocessing}).
Second, we use these graphs to pre-train a masked graph autoencoder equipped with adaptive channel-mixing message passing (Section~\ref{sec:pretraining}).
Finally, the resulting encoder provides representations that can be used directly at the cell level or aggregated to characterize tissue regions and whole slides (Section~\ref{sec:inference}).
We consider two variants of the framework: GrapHist-S, which models spatial relationships between cells through distance-weighted edges, and GrapHist-E, which additionally incorporates descriptors of the intercellular tissue and relational properties of neighboring cells.

\subsection{From histopathology images to cell graphs}
\label{sec:preprocessing}

\subsubsection{Cell segmentation}

Whole-slide images are gigapixel-resolution scans of digitized biopsies, typically stained with Hematoxylin and Eosin (H\&E).
As in~\citet{pati2022hierarchical}, we transform these images into cell graphs to explicitly capture the cellular composition and spatial arrangement, following the procedure detailed in Figure~\ref{fig:preprocessing}.
Following prior cell-graph approaches~\citep{ogut2025vision}, we detect nuclei at the patch level using StarDist~\citep{schmidt2018}, a modern, lightweight, high-performing model with a U-Net backbone, and then treat each detected nucleus as a proxy for an individual cell.
This step abstracts the image from raw pixels to its cellular constituents, focusing learning on cell morphology and spatial organization.

\subsubsection{Graph construction}

We then model the image as an undirected cell graph $\mathcal{G} = (\mathcal{V}, \mathcal{E})$, where $\mathcal{V}$ denotes its set of $n$ nodes and $\mathcal{E}$ its set of edges.
The nodes $\{v_i\}_{i \in \left[n \right]} \in \mathcal{V}$ correspond to cells and are endowed with features $\rmX = (\vx_i) \in \R^{n \times 96}$, detailed in Appendix Table~\ref{tab:node_features}, describing cells' morphology, texture, and color intensities, which are known to be discriminant across cell and tissue types~\citep{zhao2023single,fournier2026prognostic}.

Spatial connectivity is defined through Delaunay triangulation of the cell centroids~\citep{elshakhs2024comprehensive, zhou2019cgc}.
To restrict message passing to local cellular neighborhoods, we remove edges connecting centroids separated by more than 100 $\mu$m~\citep{karl1997pnas}.
Each remaining edge $(v_i, v_j) \in \mathcal{E}$ is associated with the centroid-to-centroid distance $a_{ij} \in (0, 100)$, yielding a \emph{sparse} weighted adjacency matrix $\rmA = (a_{ij}) \in \R^{n \times n}$.
The resulting attributed graph, denoted $(\rmX, \rmA)$, constitutes the input to GrapHist-S.

Our edge-enhanced variant, GrapHist-E, retains information from the image context that a purely cell-centric representation discards.
For each connected pair of cells, we define an oriented image region spanning their centroids and extract a $74$-dimensional descriptor $\ve_{ij}$ characterizing the intercellular image context and the relationship between the two cells.
The descriptors comprise intensity, texture, and morphological properties of the intercellular region together with spatial and appearance-based relational features of the connected cells.
Appendix Table~\ref{tab:edge_features} reports the complete set.
We collect these descriptors into a sparse edge-feature tensor $\rmE \in \R^{n \times n \times 74}$, where $\rmE_{ij,:}=\ve_{ij}$ for $(v_i,v_j)\in\mathcal{E}$ and $\mathbf{0}$ otherwise.
Thus, $\rmE$ exhibits a similar sparsity to $\rmA$, and GrapHist-E receives $(\rmX,\rmA,\rmE)$.

Finally, we compute all features once offline using a GPU-compatible extraction pipeline, enabling parallelized pre-processing.
Table~\ref{tab:preproc_runtime} reports the average pre-processing runtime.

\subsection{Masked graph autoencoding with adaptive channel mixing}
\label{sec:pretraining}

\begin{figure}[t]
    \centering
    \includegraphics[width=\linewidth]{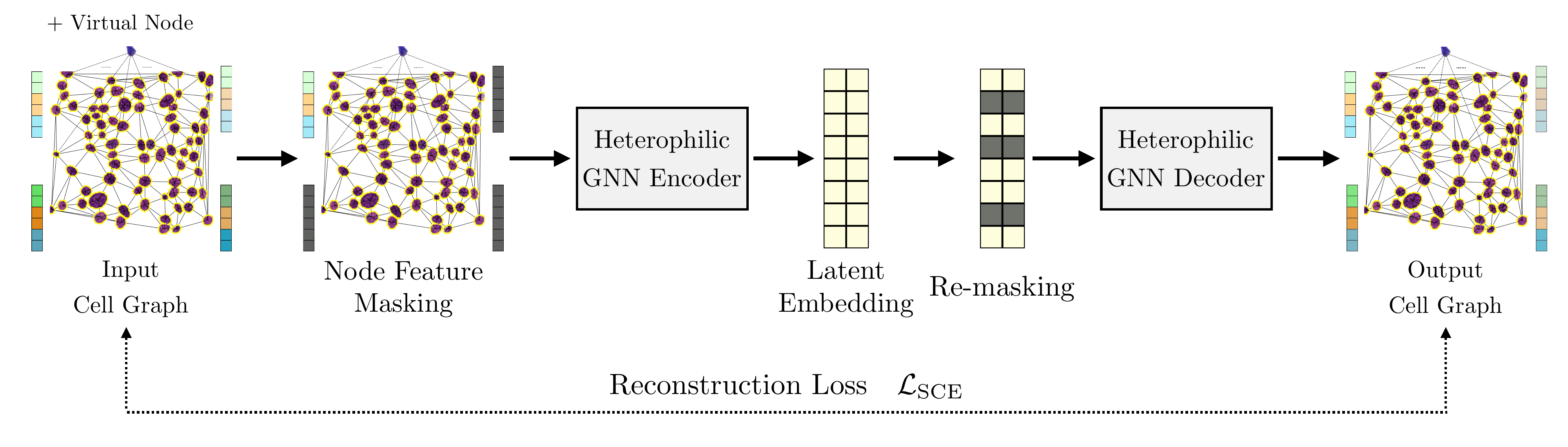}
    \caption{\textbf{Pre-training of GrapHist.} We add a virtual node that is connected to all other nodes in the input cell graph and adopt a masked autoencoding strategy, where a subset of input node features is randomly masked. A GNN-based encoder–decoder architecture is then trained to recover the original node features, decoding from a re-masked version of the latent embeddings.}
    \label{fig:main}
\end{figure}

The core principle of the GrapHist framework is graph-based self-supervision that combines masked node feature reconstruction with adaptive message passing to learn representations of spatial cellular organization.
In this section, we first describe the pre-training objective and the encoder architecture, which we instantiate in two variants, GrapHist-S and GrapHist-E, and then the mechanisms that improve its expressivity.
Figure~\ref{fig:main} illustrates the overall framework.

\subsubsection{Self-supervised objective}
\label{sec:objective}

GrapHist leverages the GraphMAE framework~\citep{hou2022graphmae}, where the features of a randomly chosen subset of nodes $\tilde{\mathcal{V}} \subset \mathcal{V}$ are masked in the input graph.
This masking procedure comes down to replacing observed node features with either a learnable vector or a randomly sampled node feature from the graph.
A GNN-based encoder then learns latent embeddings from this corrupted graph.
During decoding, the previously masked nodes are re-masked, and a GNN-based decoder is trained to reconstruct their original features $\hat{\vx}_i$ from the learned embeddings $\vh_i$.
Both the encoder and decoder use the architecture from Section~\ref{sec:backbone}.
The objective is to minimize the scaled cosine error defined as follows, where the scaling factor $\gamma \geq 1$ is adopted from~\citet{hou2022graphmae} to down-weight the contribution of easy samples:
\begin{equation}
\label{eq:sce}
    \mathcal{L}_{\text{SCE}} = \frac{1}{|\tilde{\mathcal{V}}|} \sum_{v_i \in \tilde{\mathcal{V}}} \left(1 - \frac{{\vx}_i^\top \hat{\vx}_i}{\|{\vx}_i\| \cdot \|\hat{\vx}_i\|}\right)^\gamma. \\
\end{equation}
While GrapHist-S showed no evidence of dimensional collapse and was therefore trained with the reconstruction objective alone, we observed reduced embedding dimensionality when introducing the richer edge representation of GrapHist-E.
We consequently augment the GrapHist-E objective with the variance and covariance regularization terms of VICReg~\citep{bardes2022vicreg} to mitigate this collapse while preserving the same reconstruction objective.
We quantify this effect and the impact of the regularization in Appendix~\ref{sec:pre_train_details}.

\subsubsection{Encoder and decoder architecture}
\label{sec:backbone}

\paragraph{Cell-based message passing.}
In our framework, both the encoder and decoder are message-passing GNNs that operate on the input cell graphs.
Our two proposed variants, namely GrapHist-S and GrapHist-E, differ in how the messages are computed from a cell's neighborhood.
Specifically, we denote $\rmH^{(k)}$ as the node feature matrix at layer $k$, with $\rmH^{(0)} = \rmX$, and define the message $\vm_i^{(k-1)}$ as a function of the neighborhood $\mathcal{N}(i)$ of a cell $v_i$:
\begin{equation}
\label{eq:aggregation}
\vm_i^{(k-1)} =
\frac{1}{[\rmD]_{ii}} \sum_{j \in \mathcal{N}(i)} a_{ij}
\begin{cases}
    \vh_j^{(k-1)} & \text{(GrapHist-S)}, \\[1ex]
    \operatorname{ReLU}\!\left( \vh_j^{(k-1)} + \tilde{\ve}_{ij} \right) & \text{(GrapHist-E)},
\end{cases}
\qquad [\rmD]_{ii} = \sum_{j \in \mathcal{N}(i)} a_{ij}.
\end{equation}
In other words, GrapHist-S recovers the random-walk aggregation $\rmA_{rw}\rmH^{(k-1)}$ with $\rmA_{rw} = \rmD^{-1}\rmA$, whereas GrapHist-E also accounts for the properties of the extracellular matrix.
Particularly, it adopts the mechanism in GINEConv~\citep{hu2020strategies}, where $\tilde{\ve}_{ij} = \rmW_e \ve_{ij}$ and $\rmW_e \in \R^{d \times 74}$ projects the edge features of Section~\ref{sec:preprocessing} into a $d$-dimensional space.
This projection is applied once, and its output is reused at every layer, leaving the per-layer cost unchanged.

\paragraph{Capturing tumor heterogeneity.}
In contrast to the homophilic GNNs used in vanilla GraphMAE, we combine the resulting messages through the Adaptive Channel Mixing (ACM) architecture~\citep{luan2022revisiting}, which better suits the inherent heterogeneity of tumor microenvironments.
At each layer, the ACM block processes the graph through three channels: low-pass, high-pass, and identity.
All are constructed from the message matrix $\rmM^{(k-1)} = (\vm_i^{(k-1)})$ of Eq.~\ref{eq:aggregation} and combined through mixing coefficients.
This yields the update rule below for layer $k$:
\begin{equation}
\label{eq:acm}
\rmH^{(k)} = \sum_{c \in \{L,H,I\}} \operatorname{diag}\!\big(\bm{\alpha}_c^{(k)}\big)\, \rmH_c^{(k)}, \qquad
\begin{cases}
    \rmH_L^{(k)} = \text{MLP}_L^{(k)}\!\left( \tfrac{1}{2}\big(\rmH^{(k-1)} + \rmM^{(k-1)}\big) \right), \\[1.5ex]
    \rmH_H^{(k)} = \text{MLP}_H^{(k)}\!\left( \tfrac{1}{2}\big(\rmH^{(k-1)} - \rmM^{(k-1)}\big) \right), \\[1.5ex]
    \rmH_I^{(k)} = \text{MLP}_I^{(k)}\!\left( \rmH^{(k-1)} \right),
\end{cases}
\end{equation}
where each node receives its own convex combination of the three channels, driven by the mixing coefficients $\bm{\alpha}_c^{(k)} \in \R^{n}$, which form the columns of $\big[\bm{\alpha}_L^{(k)} \,\|\, \bm{\alpha}_H^{(k)} \,\|\, \bm{\alpha}_I^{(k)}\big] = \operatorname{Softmax}\!\left( \tfrac{1}{T}\big[\hat{\bm{\alpha}}_L^{(k)} \,\|\, \hat{\bm{\alpha}}_H^{(k)} \,\|\, \hat{\bm{\alpha}}_I^{(k)}\big]\, \rmW_{\text{mix}}^{(k)} \right)$, with per-channel scores $\hat{\bm{\alpha}}_c^{(k)} = \operatorname{Sigmoid}\!\big(\rmH_c^{(k)}\rmW_c^{(k)}\big) \in \R^{n}$, weights $\rmW_c^{(k)} \in \R^{d \times 1}$, temperature $T \in \R^+$, and a mixing matrix $\rmW_{\text{mix}}^{(k)} \in \R^{3\times3}$ applied across channels.

\paragraph{Improving expressivity.}
We enhance the model's representational capacity with two mechanisms.
First, to better encode long-range dependencies within graphs, we introduce a virtual node connected to all nodes in each graph~\citep{gilmer2017neural, southern2025understanding}.
We initialize the virtual node's features to zero and set its edge weights to the mean of all edge weights in the pre-training dataset.
Second, to improve information flow across layers and prevent oversmoothing, we employ a jumping knowledge strategy~\citep{xu2018representation}.
This mechanism acts as a residual connection by concatenating the outputs from each layer, followed by a linear projection to a fixed dimension $d$, yielding final node embeddings $\rmH \in \R^{n \times d}$.

\paragraph{Computational efficiency.}
Overall, our message-passing architecture enables encoding tissues with computational complexity that is \emph{linear} in the number of nodes and edges.
Our sparse cell graph construction, based on Delaunay triangulation and distance-based edge pruning, keeps the number of edges proportional to those of nodes, yielding an overall encoding complexity that is \emph{linear} in the number of cells.
This positions GrapHist as a memory-efficient framework, unlike vision transformer models commonly used in digital pathology, whose complexity is \emph{quadratic} in the number of tokens, which is roughly four times the number of cells at a standard magnification of 20$\times$.
This scalability is particularly relevant for pathology slides, which can encompass millions of cells.

\begin{algorithm}[t]
\small
\caption{The GrapHist Framework}
\label{alg:graphist}
\vspace{0.5em}
\KwIn{Histopathology dataset $\mathcal{D}_I$.}
\KwOut{Pre-trained encoder $\mathcal{F}_{\theta}$, cell, region, and slide embeddings.}
\vspace{0.5em}
\textbf{Pre-processing (Section~\ref{sec:preprocessing}):}\\
\ForEach{image $I \in \mathcal{D}_I$}{
Perform cell segmentation on $I$ with StarDist~\citep{schmidt2018}. \\
Extract node-level features as detailed in Table~\ref{tab:node_features}. \\
Construct edges using Delaunay triangulation to obtain $\mathcal{G} = (\mathcal{V}, \mathcal{E})$. \\
Extract edge-level features as detailed in Table~\ref{tab:edge_features}. \\
}
Store the graph-based dataset in $\mathcal{D}_G$. \\
\vspace{0.5em}
\textbf{Pre-training (Section~\ref{sec:pretraining}):}\\
\For{epoch $= 1$ \KwTo 100}{
    Sample a batch of graphs $\{\mathcal{G}\} \subset \mathcal{D}_G$. \\
    Enrich them with virtual nodes. \\
    \ForEach{graph $\mathcal{G}$}{
        Sample a subset of nodes $\tilde{\mathcal{V}} \subset \mathcal{V}$. \\
        Mask their features with probability $r_m$. \\
        Randomly replace masked features with probability $r_r$. \\
        Encode the graph using a heterophilic GNN (Eq.~\ref{eq:acm}). \\
        Re-mask the same nodes in $\tilde{\mathcal{V}}$. \\
        Decode the graph with a single-layer heterophilic GNN (Eq.~\ref{eq:acm}). \\
        Compute the reconstruction loss $\mathcal{L}_{\text{SCE}}$ (Eq.~\ref{eq:sce}). \\
        }
    Update encoder and decoder parameters via backpropagation. \\
}
\Return{$\mathcal{F}_{\theta}$.} \\
\vspace{0.5em}
\textbf{Inference (Section~\ref{sec:inference}):}\\
\ForEach{graph $\mathcal{G}$}{
    Compute cell embeddings $\rmH = \mathcal{F}_{\theta}(\mathcal{G}) \in \R^{n \times d}$. \\
    \textbf{Cell level:} Use the rows $\vh_i$ of $\rmH$ directly. \\
    \textbf{Region level:} Mean pool into $\vh = \tfrac{1}{n}\sum_{i} \vh_i \in \R^{d}$. \\
    \textbf{Slide level:} Aggregate region embeddings with attention-based MIL. \\
}
\end{algorithm}

\subsection{Deriving cell-, region-, and slide-level embeddings}
\label{sec:inference}

Following Algorithm~\ref{alg:graphist}, we pre-train our models on a large dataset of cell graphs to obtain context-aware cell-level embeddings.
This enables our framework to operate across biological scales.
In other words, we can directly utilize the produced node embeddings $\rmH \in \R^{n \times d}$ for cell-level tasks, such as cell type identification.
We can also aggregate them into a region-level embedding $\vh \in \R^d$, defined as the mean of cell embeddings, to address localized tumor subtyping tasks.
Finally, we can further aggregate the derived region embeddings into slide-level embeddings using standard attention-based multiple instance learning (MIL) approaches, commonly used in digital pathology~\citep{gadermayr2024multiple}.
In particular, we consider the best-performing method among three state-of-the-art attention-based MIL approaches: ABMIL~\citep{ilse2018attention}, additive ABMIL~\citep{javed2022additive}, and conjunctive ABMIL~\citep{early2024inherently}, which mainly differ in how they couple the attention mechanism with a classifier.
Formally, they compute the slide logits $\vs_i$:
\begin{equation*}
    \underbrace{
        \vs_i = g \left( \sum_{j=1}^{K_i} \lambda_{ij} \vh_{ij} \right),
    }_{\text{(ABMIL)}} \quad
    \underbrace{
        \vs_i = \sum_{j=1}^{K_i} g(\lambda_{ij} \vh_{ij}),
    }_{\text{(add-ABMIL)}} \quad
    \underbrace{
        \vs_i = \sum_{j=1}^{K_i} \lambda_{ij}\, g( \vh_{ij}),
    }_{\text{(conj-ABMIL)}}
\end{equation*}
\noindent where $\{ \vh_{ij} \}_{j \in \left[K_i \right]}$ are the $K_i$ region embeddings of the slide $i$, $g(\cdot)$ is a classifier, and each attention coefficient $\lambda_{ij}$ is defined as $\exp(\vw^\top \tanh(\rmV\vh_{ij}^\top)) /$ $\sum_{k=1}^{K_i} \exp(\vw^\top \tanh(\rmV\vh_{ik}^\top))$ with $\vw$ and $\rmV$ as learnable parameters.
\section{Experiments}
\label{sec:experiments}

\subsection{Datasets}
\label{sec:datasets}

For self-supervised pre-training, we utilize a dataset of 998 H\&E-stained breast cancer WSIs from the TCGA database~\citep{weinstein2013cancer}, comprising about 11 million 224$\times$224-pixel patches.
This dataset also serves as an in-domain (ID) evaluation, through the classification of infiltrating ductal carcinoma and lobular carcinoma across different grades, as well as survival analysis.
To assess out-of-domain (OOD) generalization, we further evaluate on five datasets spanning breast and other tissue types and two biological scales: BACH~\citep{aresta2019bach}, BRACS~\citep{brancati2022bracs}, and BreakHis~\citep{spanhol2015dataset} at the region level, and PanNuke~\citep{gamper2020pannukedatasetextensioninsights} and NuCLS~\citep{Amgad_2022} at the cell level.
NuCLS includes TCGA-BRCA slides, sharing the source of our ID dataset, and therefore does not constitute a strictly OOD evaluation.
Appendix Table~\ref{tab:dataset_stats} summarizes their statistics and associated tasks, with further details in Appendix~\ref{sec:dataset_details}.
Two of these datasets support several evaluation settings.
On PanNuke, we report results at both 20$\times$ and 40$\times$ magnification to assess how representation quality depends on image resolution.
In each case, we evaluate all pan-cancer tissues as well as the breast subset alone, which is closer to our pre-training distribution.
On NuCLS, we consider both the fine-grained \emph{main} labels and the coarser \emph{super} grouping.
We convert all image datasets into their graph counterparts following Section~\ref{sec:preprocessing}.

\subsection{Baselines}

\subsubsection{Self-supervised methods}

\paragraph{Vision baselines.}
To quantify the benefits of incorporating prior biological knowledge through cell graphs, we compare GrapHist against two widely used self-supervised vision models that underpin many contemporary pathology foundation models.
DINOv2~\citep{oquab2023dinov2}, which builds on self-distillation using a student-teacher pair of vision transformers, is now one of the most widely used approaches~\citep{campanella2024computational}, while earlier masked autoencoding frameworks such as MAE~\citep{he2022masked} remain competitive and are often preferred for generative tasks~\citep{kraus2024masked}.

\paragraph{Pre-training protocol.}
To ensure a fair comparison, we pre-train DINOv2 and MAE \textit{from scratch} on patch-level images from TCGA-BRCA, whereas GrapHist variants are pre-trained on graphs generated from the same patches, all for 100 epochs with a batch size of 2048.
For the vision models, we use a ViT-S/14 backbone~\citep{dosovitskiy2020image} with the standard hyperparameters provided by the authors, as they are well validated across data scales and architectures.
For GrapHist, the lack of large-scale benchmarks for the GraphMAE framework instead requires validating its key hyperparameters following the guidelines of~\citet{hou2022graphmae}: the embedding dimension in $\{512, 768, 1024\}$, the masking ratio $r_m$ in $\{0.50, 0.75\}$, and the replacement ratio $r_r$ in $\{0.00, 0.10\}$.
We set encoder and decoder depths to five and one, respectively, again following~\citet{hou2022graphmae} for large datasets.
We validate these choices in Appendix~\ref{sec:sensitivity} with a sensitivity analysis for GrapHist-S and adopt the same parameters for GrapHist-E.

\paragraph{Evaluation.}
For region- and slide-level tasks, we use the CLS token of the vision transformers and the mean of GrapHist's cell embeddings.
For cell-level tasks, GrapHist directly provides cell embeddings through its node representations, whereas the vision baselines operate on fixed-grid tokens that are misaligned with irregular cellular boundaries.
We therefore obtain their cell embeddings as convex combinations of the tokens spatially overlapping each nucleus, weighted by the corresponding segmentation masks.
We keep all representations frozen and evaluate them with linear probing for cell-level tasks and with the attention-based MIL aggregators of Section~\ref{sec:inference} for region- and slide-level benchmarks.
Accordingly, we train one MIL head per fold of a stratified five-fold cross-validation on the training set and report the mean $\pm$ standard deviation of test macro F1 across the five resulting models on the held-out test set.
For cell-level tasks, we train and test the probe on each official split of PanNuke (three folds) and NuCLS (five folds), and report the mean $\pm$ standard deviation across folds.
Further details are deferred to Appendix~\ref{sec:setup_ssl}.

\subsubsection{Supervised methods}

\paragraph{Graph baselines.}
To isolate the benefits of self-supervised pre-training and of our hand-crafted features, we further compare against two fully supervised graph baselines, both sharing the ACM backbone but trained end-to-end on the downstream labels of each dataset.
The first, ACM-bio, operates on exactly the same attributed graphs as GrapHist-S, so that any difference in performance is attributable to the training regime rather than to the input representation.
The second, ACM-UNI, instead replaces our hand-crafted cell descriptors with deep features from UNI~\citep{chen2024towards}, a well-established pathology foundation model, obtained by resizing each segmented cell (typically ${\sim}20\times20$ pixels) to 224$\times$224 pixels before encoding.
Together, these comparisons disentangle the contributions of our descriptors and large-scale pre-training.
Since both baselines overfit under the hyperparameter ranges of GrapHist-S, we validate them over separate, task-dependent ranges detailed in Appendix~\ref{sec:setup_sl}.
Given the overfitting observed in these baselines, we did not extend them to the more parameterized GrapHist-E variant.
\section{Results}

We evaluate GrapHist-S and GrapHist-E along two dimensions.
First, we examine whether a single pre-trained encoder transfers across cell-, region-, and slide-level tasks.
Second, we assess the computational efficiency of the proposed graph-based representation and its ability to scale to graphs substantially larger than those observed during pre-training.

\begin{table*}[t]
\caption{Test macro F1 scores (\%) on slide- and region-level tumor subtyping, reported as mean $\pm$ standard deviation over a five-fold cross-validation on the training set, each evaluated on the same held-out test set. Best and second-best results in each row are highlighted in bold and underlined, respectively.}
\label{tab:tumor_subtyping}
\centering
\small
\setlength{\tabcolsep}{5pt}
\resizebox{\textwidth}{!}{%
\begin{tabular}{lcccccc}
\toprule
&
\multicolumn{2}{c}{\textbf{Supervised graph models}}
&
\multicolumn{2}{c}{\textbf{Self-supervised vision models}}
&
\multicolumn{2}{c}{\textbf{Ours}}
\\
\cmidrule(lr){2-3}
\cmidrule(lr){4-5}
\cmidrule(lr){6-7}
\textbf{Dataset}
& ACM-bio
& ACM-UNI
& DINOv2
& MAE
& GrapHist-S
& GrapHist-E
\\
\midrule

\multicolumn{7}{l}{\textbf{Slide-level}} \\[2pt]

TCGA-BRCA
& OOM
& OOM
& 62.43 {\scriptsize $\pm$ 1.75}
& 66.72 {\scriptsize $\pm$ 1.27}
& \textbf{72.25 {\scriptsize $\pm$ 1.40}}
& \underline{72.06 {\scriptsize $\pm$ 2.50}}
\\

\midrule
\multicolumn{7}{l}{\textbf{Region-level}} \\[2pt]

BACH
& 38.37 {\scriptsize $\pm$ 2.16}
& 29.03 {\scriptsize $\pm$ 9.92}
& 59.25 {\scriptsize $\pm$ 3.34}
& 57.94 {\scriptsize $\pm$ 3.91}
& \textbf{69.16 {\scriptsize $\pm$ 3.37}}
& \underline{68.98 {\scriptsize $\pm$ 3.59}}
\\

BRACS
& 21.55 {\scriptsize $\pm$ 2.62}
& 19.95 {\scriptsize $\pm$ 3.27}
& 52.68 {\scriptsize $\pm$ 1.78}
& 56.33 {\scriptsize $\pm$ 0.76}
& \underline{60.30 {\scriptsize $\pm$ 0.46}}
& \textbf{69.00 {\scriptsize $\pm$ 1.30}}
\\

BreakHis
& 55.82 {\scriptsize $\pm$ 10.33}
& 76.27 {\scriptsize $\pm$ 9.23}
& 83.44 {\scriptsize $\pm$ 2.25}
& 87.74 {\scriptsize $\pm$ 1.66}
& \underline{89.37 {\scriptsize $\pm$ 1.94}}
& \textbf{95.53 {\scriptsize $\pm$ 2.44}}
\\

\bottomrule
\end{tabular}%
}
\end{table*}

\subsection{Transfer across biological scales}
\label{sec:cross_scale_transfer}

\subsubsection{Tumor subtyping}

Table~\ref{tab:tumor_subtyping} summarizes tumor-subtyping performance across slide- and region-level datasets.
Our methods achieve the strongest performance on all four tumor-subtyping benchmarks.
On the in-domain TCGA-BRCA datasets, GrapHist-S achieves $72.25\%$ macro F1, exceeding the strongest vision baseline by $5.53$ percentage points.
Here, the supervised baselines require storing all patch-level graphs of a slide simultaneously for backpropagation, which leads to prohibitive memory costs (OOM).
On the out-of-domain region-level datasets, our models surpass the baselines by margins of up to $9.9$ percentage points.
On the contrary, the fully-supervised graph methods perform significantly worse on these tasks, potentially due to their tendency to overfit to limited downstream labels.
Finally, the two GrapHist configurations reveal a task-dependent contribution of intercellular context.
GrapHist-S and GrapHist-E perform comparably on TCGA-BRCA and BACH, whereas adding intercellular edge descriptors improves macro F1 by \(8.70\) points on BRACS and \(6.16\) points on BreakHis.
This suggests that cellular morphology and spatial organization alone suffice for some tasks, while richer intercellular information benefits others.
The particularly large gain on BRACS is consistent with its diagnostic setting, where several lesion categories differ not only in cellular atypia but also in ductal architecture, lesion extent, and stromal invasion~\citep{brancati2022bracs}.

\begin{figure}[htbp]
    \centering
    \begin{subfigure}[b]{0.24\linewidth}
        \centering
        \includegraphics[width=\linewidth]{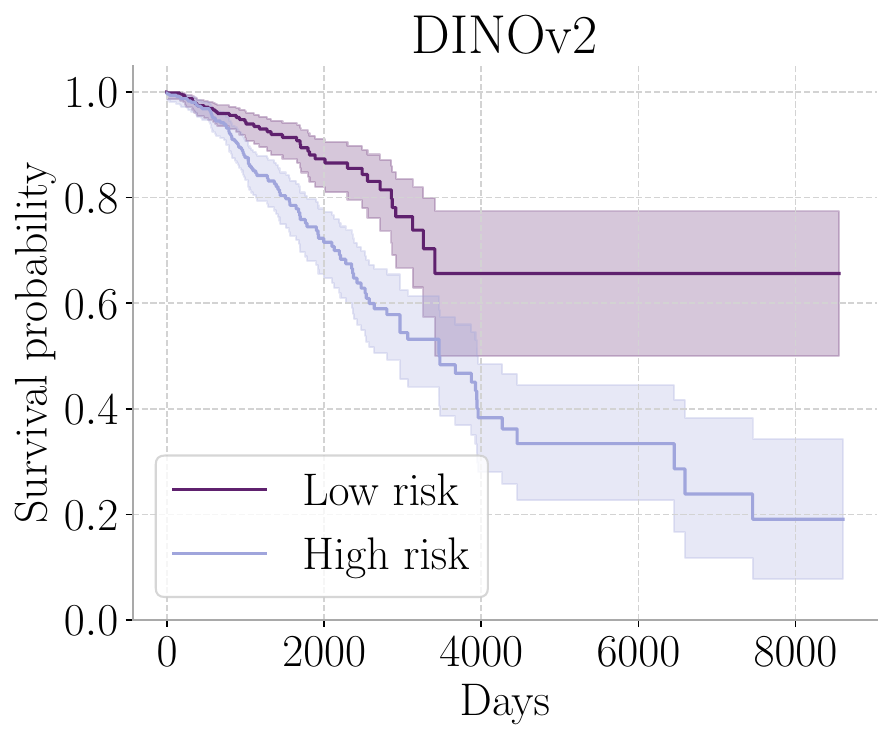}
    \end{subfigure}
    \begin{subfigure}[b]{0.24\linewidth}
        \centering
        \includegraphics[width=\linewidth]{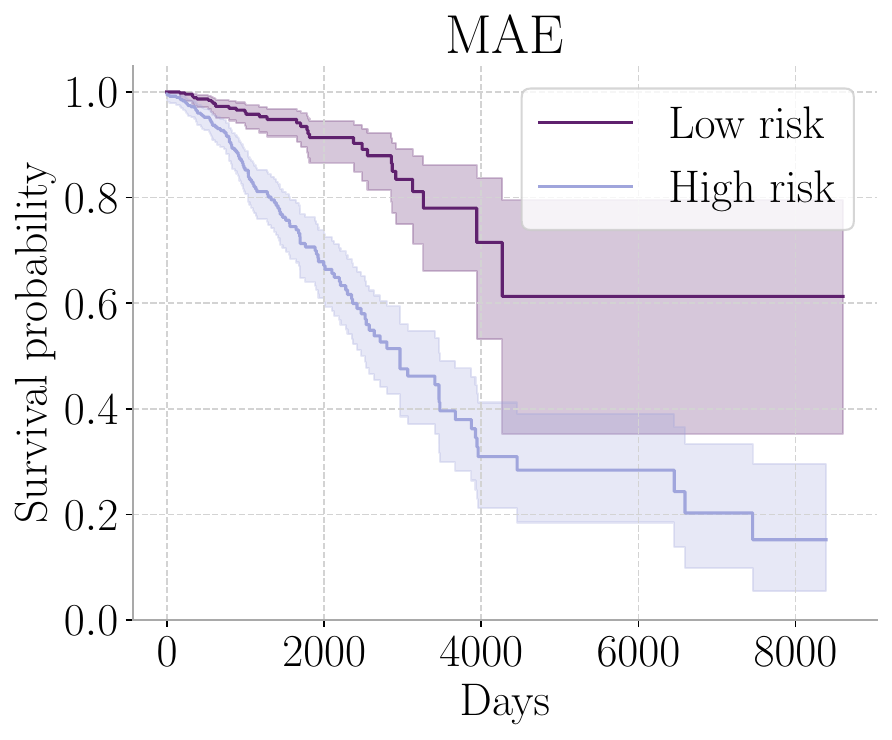}
    \end{subfigure}
    \begin{subfigure}[b]{0.24\linewidth}
        \centering
        \includegraphics[width=\linewidth]{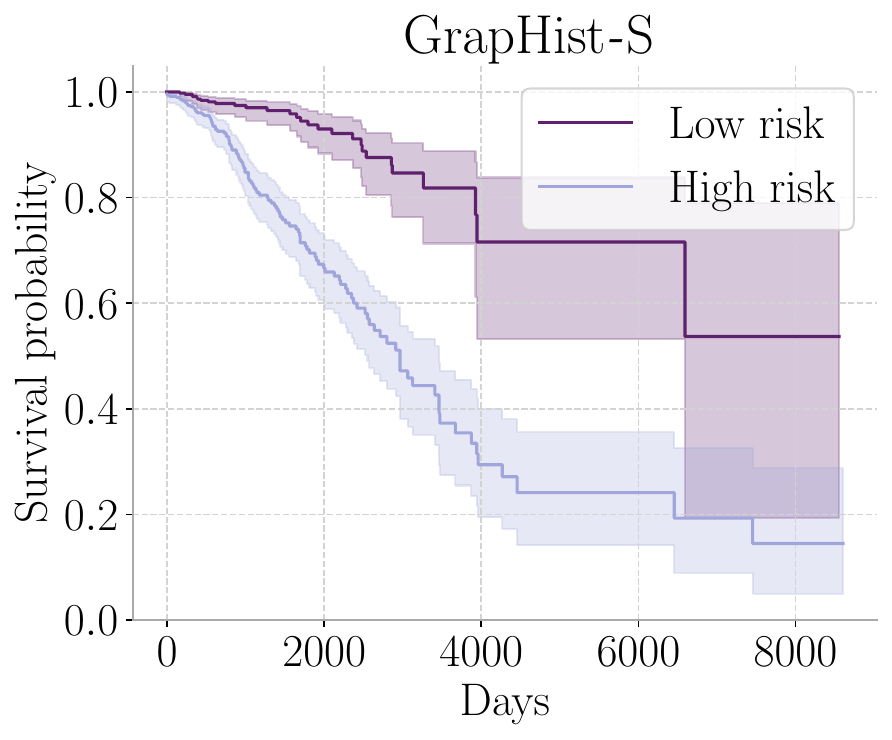}
    \end{subfigure}
    \begin{subfigure}[b]{0.24\linewidth}
        \centering
        \includegraphics[width=\linewidth]{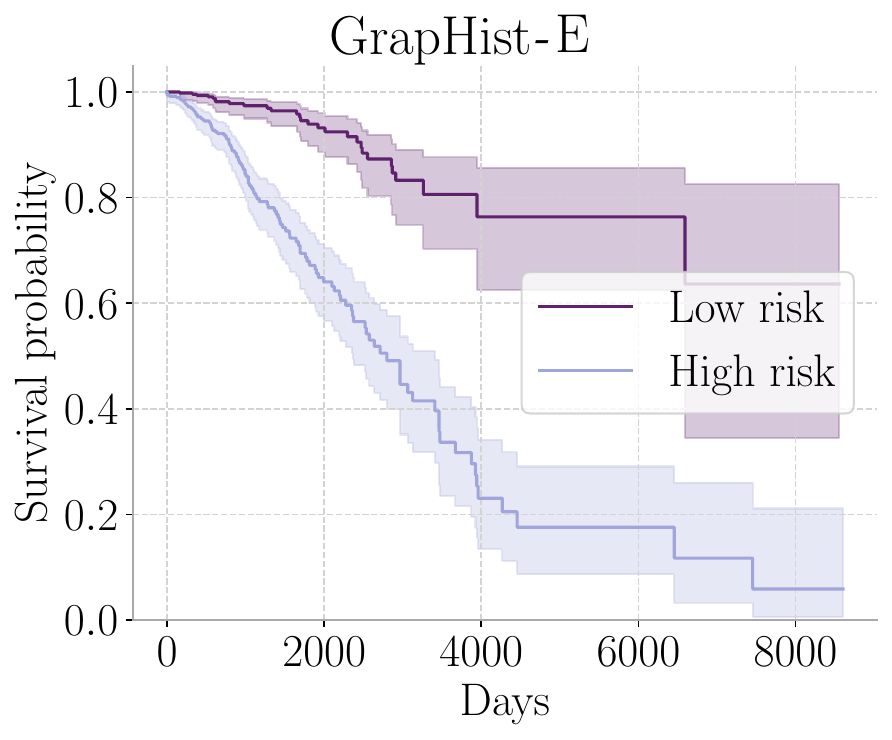}
    \end{subfigure}
    \caption{Kaplan-Meier curves with Cox risk groups.}
    \label{fig:kaplan_meier}
\end{figure}

\subsubsection{Survival analysis}

\setlength{\intextsep}{0pt}%
\begin{wraptable}{r}{0.45\textwidth}
    \caption{Survival analysis on TCGA-BRCA.}
    \label{tab:survival_analysis}
    \centering
    \begin{tabular}{lcc}
        \toprule
        \textbf{Model} & \textbf{C-index} & \textbf{p-value} \\
        \midrule
        DINOv2    & 0.63 & 9.80 $\times 10^{-6}$ \\
        MAE       & 0.72 & 1.54 $\times 10^{-13}$ \\
        GrapHist-S & 0.76 & 1.29 $\times 10^{-15}$ \\
        GrapHist-E & \textbf{0.79} & \textbf{2.16 $\times 10^{-20}$} \\
        \bottomrule
    \end{tabular}
\vspace{-2pt}
\end{wraptable}

We further evaluate the prognostic value of the learned embeddings on TCGA-BRCA by performing survival analysis, i.e., Kaplan-Meier~\citep{kaplan1958nonparametric} and Cox analysis~\citep{cox1972regression}.
Following~\citet{wulczyn2020deep}, for each patient, we first compute a patient-level embedding by averaging their WSI embeddings, themselves defined as the mean of patch embeddings obtained directly from pre-trained models.
Overall survival time is defined as days-to-death for deceased patients and days-to-last-follow-up for censored patients, with the event indicator set to 1 for death and 0 otherwise.
As is typical in survival analysis, we then fit a penalized Cox proportional hazards model using the patient embeddings as covariates, compute risk scores, and stratify patients into high- and low-risk groups based on the median risk score.
Group separation is assessed using Kaplan–Meier curves and log-rank tests.
As shown in Figure~\ref{fig:kaplan_meier}, GrapHist-E achieves more pronounced risk-group separation compared to vision-based baselines.
We report the concordance index (C-index), which measures the model's ability to correctly rank patient risk (higher is better), and the p-value, which quantifies the statistical significance of the stratification.
As seen in Table~\ref{tab:survival_analysis}, GrapHist-E achieves the strongest performance, followed by GrapHist-S, and then MAE, while DINOv2 is consistently weaker, indicating the structure-aware benefit of our approach.
\newline

\begin{table}[t]
\caption{Test macro F1 scores (\%) on cell phenotyping, reported as mean $\pm$ standard deviation across the official folds of each dataset (three for PanNuke, five for NuCLS). Best and second-best results in each row are highlighted in bold and underlined, respectively.}
\label{tab:cell_phenotyping}
\centering
\resizebox{\textwidth}{!}{%
\begin{tabular}{llcccccc}
\toprule
& & \multicolumn{2}{c}{\textbf{Supervised graph models}}
& \multicolumn{2}{c}{\textbf{Self-supervised vision models}}
& \multicolumn{2}{c}{\textbf{Ours}} \\
\cmidrule(lr){3-4}
\cmidrule(lr){5-6}
\cmidrule(lr){7-8}
\textbf{Dataset} & \textbf{Setting}
& ACM-bio & ACM-UNI & DINOv2 & MAE & GrapHist-S & GrapHist-E \\
\midrule
\multirow{2}{*}{PanNuke 20$\times$}
& Breast
& 56.61 \scriptsize{$\pm$ 0.31}
& \underline{58.05 \scriptsize{$\pm$ 0.03}}
& 54.82 \scriptsize{$\pm$ 0.48}
& 47.71 \scriptsize{$\pm$ 0.64}
& 55.26 \scriptsize{$\pm$ 0.30}
& \textbf{58.57 \scriptsize{$\pm$ 0.33}} \\
& PanCancer
& 62.73 \scriptsize{$\pm$ 0.02}
& \textbf{67.63 \scriptsize{$\pm$ 0.66}}
& 50.49 \scriptsize{$\pm$ 0.37}
& 54.88 \scriptsize{$\pm$ 0.55}
& 58.78 \scriptsize{$\pm$ 0.15}
& \underline{63.25 \scriptsize{$\pm$ 0.41}} \\
\midrule
\multirow{2}{*}{PanNuke 40$\times$}
& Breast
& \underline{57.06 \scriptsize{$\pm$ 0.33}}
& 56.81 \scriptsize{$\pm$ 0.18}
& 53.86 \scriptsize{$\pm$ 0.54}
& 47.54 \scriptsize{$\pm$ 0.58}
& 56.43 \scriptsize{$\pm$ 0.04}
& \textbf{58.38 \scriptsize{$\pm$ 0.18}} \\
& PanCancer
& \underline{65.80 \scriptsize{$\pm$ 0.11}}
& \textbf{66.99 \scriptsize{$\pm$ 0.80}}
& 49.27 \scriptsize{$\pm$ 0.42}
& 55.26 \scriptsize{$\pm$ 0.12}
& 59.47 \scriptsize{$\pm$ 0.11}
& 63.35 \scriptsize{$\pm$ 0.43} \\
\midrule
\multirow{2}{*}{NuCLS}
& main
& 22.13 \scriptsize{$\pm$ 1.59}
& 21.68 \scriptsize{$\pm$ 1.03}
& 21.42 \scriptsize{$\pm$ 1.49}
& 25.19 \scriptsize{$\pm$ 3.14}
& \underline{26.57 \scriptsize{$\pm$ 2.16}}
& \textbf{27.09 \scriptsize{$\pm$ 2.57}} \\
& super
& 37.35 \scriptsize{$\pm$ 1.30}
& 39.46 \scriptsize{$\pm$ 1.33}
& 41.17 \scriptsize{$\pm$ 2.26}
& 45.31 \scriptsize{$\pm$ 1.05}
& \textbf{46.24 \scriptsize{$\pm$ 3.89}}
& \underline{45.74 \scriptsize{$\pm$ 3.54}} \\
\bottomrule
\end{tabular}}
\end{table}

\subsubsection{Cell phenotyping}
\label{sec:cell_phenotyping}

We next evaluate whether the pre-trained graph encoder transfers at its native cellular resolution, using node embeddings directly for cell phenotyping.
As shown in Table~\ref{tab:cell_phenotyping}, both GrapHist configurations outperform the matched self-supervised vision baselines across all six settings, showing that the benefit of the cellular graph representation persists without region- or slide-level aggregation.
Within the GrapHist family, incorporating intercellular descriptors further improves performance across all PanNuke settings while yielding comparable results on NuCLS.

The supervised graph controls further reveal how performance changes with source--target alignment.
NuCLS is derived from TCGA-BRCA slides and therefore remains closely aligned with our pre-training distribution.
Here, GrapHist-S substantially outperforms ACM-bio despite using no downstream labels during representation learning.
The breast subset of PanNuke preserves tissue identity but introduces a distinct benchmark distribution, including differences in cohort composition, annotation protocol, and potentially image acquisition and pre-processing.
Under this moderate shift, GrapHist-S remains close to ACM-bio, while GrapHist-E exceeds it at both magnifications.
The largest shift occurs on pan-cancer PanNuke, which additionally introduces tissue types absent from our pre-training.
Here, the target-supervised models become comparatively stronger, particularly ACM-UNI.
This is consistent with ACM-UNI benefiting from two sources of alignment unavailable to frozen GrapHist variants: direct supervision on the target dataset and UNI features pretrained across multiple tissue types.
Importantly, UNI features do not consistently improve over the biological descriptors in the more closely aligned settings, suggesting that their advantage becomes most apparent as the evaluation distribution moves further from the breast-restricted pre-training domain.

\paragraph{Parameter-efficient adaptation.}
\label{sec:cell_peft}

The previous results suggest that the main limitation of frozen GrapHist-S representations emerges as the target distribution departs from the breast-restricted pre-training domain.
We therefore investigate whether limited target-domain supervision can correct this mismatch without updating the full encoder.
We augment GrapHist-S with a mixing-guided adaptation scheme inspired by AdapterGNN~\citep{li2024adaptergnn}, in which the pre-trained channel-mixing coefficients guide two bottleneck adapter branches, as detailed in  Appendix~\ref{sec:fine_tune_details}. Table~\ref{tab:results_cell_peft} shows that adaptation is most beneficial in the cross-tissue PanNuke setting, where it substantially closes the gap with the matched fully supervised ACM-bio model.
In the more closely aligned breast-only PanNuke and NuCLS settings, gains are comparatively limited and task-dependent, consistent with less distributional mismatch to correct.
Overall, these results suggest that parameter-efficient adaptation can compensate for part of the transfer gap induced by restricted pre-training diversity, while pointing toward the same scaling direction pursued by pathology foundation models, since nothing in the GrapHist-S formulation inherently limits pre-training on larger, more diverse cohorts.

\begin{table}[t]
\caption{
Cell-phenotyping macro F1 scores (\%) after parameter-efficient adaptation of GrapHist-S, reported as mean $\pm$ standard deviation across the official folds of each dataset (three for PanNuke, five for NuCLS).
$\Delta_{\mathrm{GrapHist\text{-}S}}$ and $\Delta_{\mathrm{ACM\text{-}bio}}$ denote absolute percentage-point differences relative to the frozen GrapHist-S representation and the matched fully supervised ACM-bio baseline, respectively.
}
\label{tab:results_cell_peft}
\centering
\begin{tabular}{llcrr}
\toprule
\textbf{Dataset} & \textbf{Setting} & \textbf{GrapHist-S-PEFT}
& \textbf{$\Delta_{\mathrm{GrapHist\text{-}S}}$}
& \textbf{$\Delta_{\mathrm{ACM\text{-}bio}}$} \\
\midrule

\multirow{2}{*}{PanNuke 20$\times$}
& Breast
& 57.16 {\scriptsize $\pm$ 0.29}
& $+1.90$
& $+0.55$ \\

& PanCancer
& 66.88 {\scriptsize $\pm$ 0.01}
& $+8.10$
& $+4.15$ \\

\midrule

\multirow{2}{*}{PanNuke 40$\times$}
& Breast
& 56.58 {\scriptsize $\pm$ 0.46}
& $+0.15$
& $-0.48$ \\

& PanCancer
& 65.99 {\scriptsize $\pm$ 0.52}
& $+6.52$
& $+0.19$ \\

\midrule

\multirow{2}{*}{NuCLS}
& Main
& 28.21 {\scriptsize $\pm$ 1.44}
& $+1.64$
& $+6.08$ \\

& Super
& 44.29 {\scriptsize $\pm$ 0.74}
& $-1.95$
& $+6.94$ \\

\bottomrule
\end{tabular}
\end{table}

\subsection{Efficiency and scaling trade-offs}

Representing tissue through explicit cells and their spatial relations changes both the inductive bias and the computational profile of the learning pipeline.
In particular, cellular and relational prior knowledge can be incorporated directly through node features, edge attributes, and graph structure, rather than being recovered implicitly from dense image tokens.
This abstraction, however, requires an additional image-to-graph transformation.
We therefore distinguish below between the cost of constructing the cellular representation and the subsequent cost of representation learning.
\newline

\begin{table}[htbp]
\centering
\begin{minipage}[t]{0.46\textwidth}
\centering
\caption{
Average pre-processing runtime in seconds per $224\times224$ patch on BACH.
Cell segmentation and graph construction are performed on CPU, while cell-feature extraction is performed on an NVIDIA RTX~5090 GPU.
}
\label{tab:preproc_runtime}
\begin{tabular}{lcc}
\toprule
\textbf{Stage} & \textbf{Device} & \textbf{Time} \\
\midrule
Cell segmentation       & CPU & 0.0594 \\
Cell feature extraction & GPU & 0.1550 \\
Graph construction      & CPU & 0.0003 \\
\midrule
Total preprocessing     & --  & 0.2147 \\
\bottomrule
\end{tabular}
\end{minipage}
\hfill
\begin{minipage}[t]{0.50\textwidth}
\centering
\caption{Pre-training and inference efficiency on an NVIDIA H200 GPU. The former is measured on TCGA-BRCA, while the latter uses BACH after offline graph construction.}
\label{tab:model_params}
\begin{tabular}{lccc}
\toprule
& \multicolumn{2}{c}{\textbf{Pre-training}}
& \textbf{Inference} \\
\cmidrule(lr){2-3}
\cmidrule(l){4-4}
\textbf{Model}
& \shortstack{\# Params\\($10^6$)}
& \shortstack{Time\\(h)}
& \shortstack{Peak memory\\(GB)} \\
\midrule
DINOv2     & 22.01         & 180         & 0.380 \\
MAE        & 47.58         & 350         & 0.309 \\
GrapHist-S & \textbf{9.50} & \textbf{50} & \textbf{0.142} \\
GrapHist-E & 10.53         & 89          & 0.327 \\
\bottomrule
\end{tabular}
\end{minipage}
\end{table}

\paragraph{Computational profile and performance trade-off.}

GrapHist changes the distribution of computational cost relative to image-based SSL rather than providing a uniformly cheaper pipeline.
Constructing the cellular representation adds an offline stage for nucleus segmentation and feature extraction, required both to build the pre-training corpus and to process previously unseen images.

As quantified in Table~\ref{tab:preproc_runtime}, this transformation requires on average \(0.2147\) s per \(224\times224\) patch in our implementation, with most of the cost coming from cell-feature extraction and, to a lesser extent, nucleus segmentation.
Constructing the sparse graph itself is negligible (\(0.0003\) s).
We obtained these measurements patch-wise on BACH and should therefore interpret them as a practical estimate of pre-processing overhead rather than an irreducible wall-clock cost.
In particular, the independent processing of patches is readily parallelizable, and CPU- and GPU-based stages can in principle be pipelined or distributed across workers.
Once constructed, however, these attributed graphs can be cached and reused, and learning operates on substantially smaller models with sparse message passing.
Accordingly, Table~\ref{tab:model_params} shows that, after graph construction, GrapHist-S requires fewer parameters and substantially less model-side pre-training time than the matched ViT-S baselines.
GrapHist-E occupies a different point in this trade-off: enriching the edges improves performance on several downstream tasks, but increases both training cost and inference memory relative to GrapHist-S.
Hence, we view the two models as lightweight and context-enriched variants of the same framework, rather than claiming a uniform computational advantage over image SSL.

\paragraph{Scaling beyond fixed-size patches.}
The preceding analysis considers the computational profile of the different representations at a fixed input scale.
A complementary question is how these representations behave as the spatial support increases.
In the main region- and slide-level benchmarks, we deliberately retain the conventional patching-and-MIL pipeline used by the vision baselines, so that the comparison primarily isolates the effect of the learned representation rather than introducing a different aggregation strategy.
This does not imply that vision transformers are restricted to the $224\times224$ input size used during pre-training: larger inputs remain possible, but dense self-attention becomes increasingly costly as the number of image tokens grows, whereas sparse message passing on the cell graph scales approximately linearly with the number of cells.
Thus, we progressively increase the graph extent while retaining the patch-based evaluation paradigm, before transitioning to direct whole-region inference.

\begin{wrapfigure}{r}{0.40\textwidth}
    \vspace{-1em}
    \centering
    \includegraphics[width=\linewidth]{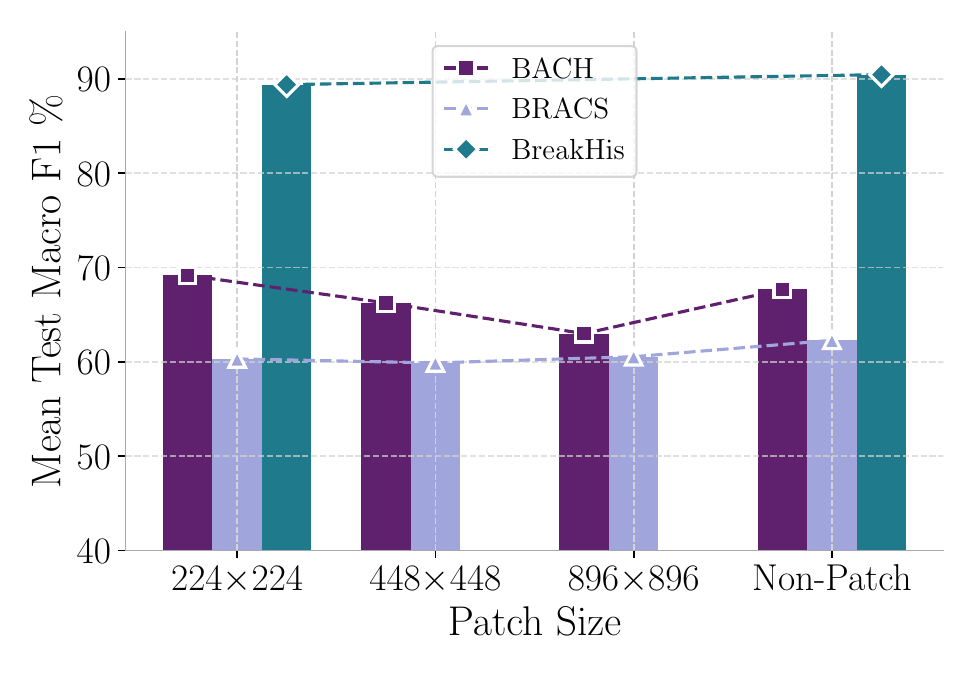}
    \caption{Region-level macro F1 as the spatial extent of the input graph increases.}
    \label{fig:patch_size}
\end{wrapfigure}
Specifically, we increase the input extent from the $224\times224$ regions used during pre-training to $448\times448$ and $896\times896$ pixels, before directly embedding complete regions of interest reaching approximately $4000\times4000$ pixels on BRACS.
At the intermediate scales, representations are evaluated through the same patchwise MIL pipeline as above, whereas complete regions are represented by a single graph whose cell embeddings are simply mean-pooled and evaluated with a linear classifier.
Performance shows no systematic degradation as the spatial extent increases, with particularly stable behavior on BRACS and BreakHis.
Most importantly, direct whole-region inference remains highly competitive despite replacing the patchwise MIL pipeline with only mean pooling over cell embeddings followed by a linear classifier.
This indicates that locally pre-trained cellular representations remain useful when composed over substantially larger spatial supports, and that region-level prediction need not rely on a fixed patch decomposition.
\section{Conclusion}

We introduced GrapHist, a self-supervised learning framework that combines masked graph autoencoding with adaptive message passing to learn transferable cell-centric representations of histopathological tissue.
Under matched source data and training exposure, GrapHist-S consistently outperforms the evaluated vision baselines across cell-, region-, and slide-level tasks, while GrapHist-E reveals task-dependent benefits from retaining richer intercellular tissue context.
Beyond predictive performance, the cell-graph abstraction provides a compact biological representation that transfers across scales and remains effective on spatial supports substantially larger than those used during pre-training.
This comes with a different computational trade-off from image SSL, introducing an explicit image-to-graph preprocessing stage but enabling subsequent learning with smaller models and sparse message passing.
Our current pre-training is restricted to breast tissue, and cross-tissue experiments indicate that broader, more diverse pre-training cohorts are an important direction for improving transfer.
GrapHist also relies on nucleus segmentation and engineered cellular and relational descriptors, motivating future work on more robust or partially learned graph construction.
More generally, the framework could be extended through stronger graph-SSL objectives, hierarchical aggregation from cells to larger tissue structures, and node- or edge-level attribution, while its cell-centric formulation provides a natural basis for incorporating complementary molecular or imaging information.
\newpage



\bibliography{reference}
\bibliographystyle{tmlr}
\newpage

\tableofcontents
\newpage

\appendix
\section{Dataset details}
\label{sec:dataset_details}

\begin{figure}[htbp]
    \centering
    \begin{subfigure}{0.49\textwidth}
        \includegraphics[width=\linewidth]{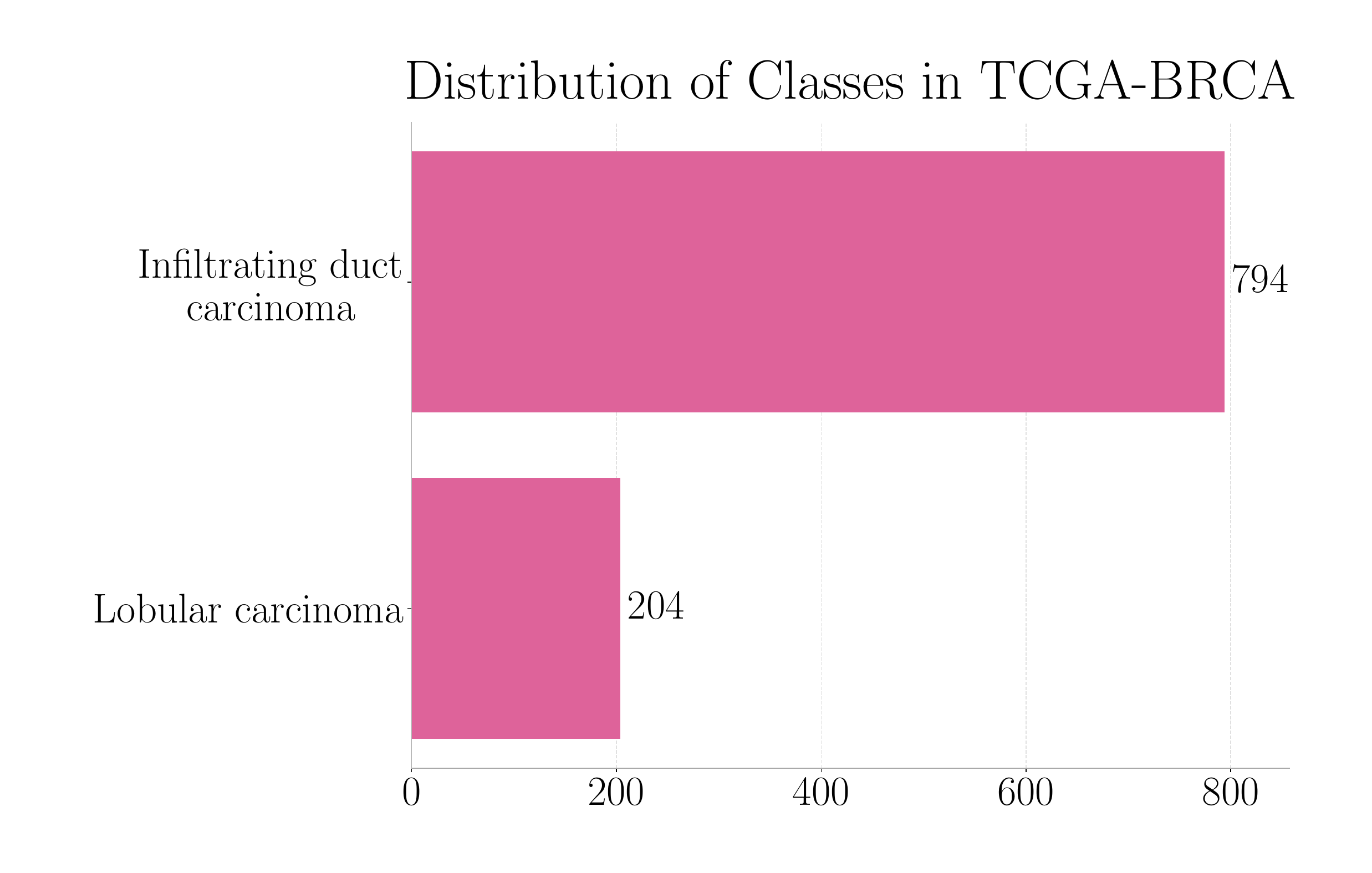}
    \end{subfigure}
    \begin{subfigure}{0.49\textwidth}
        \includegraphics[width=\linewidth]{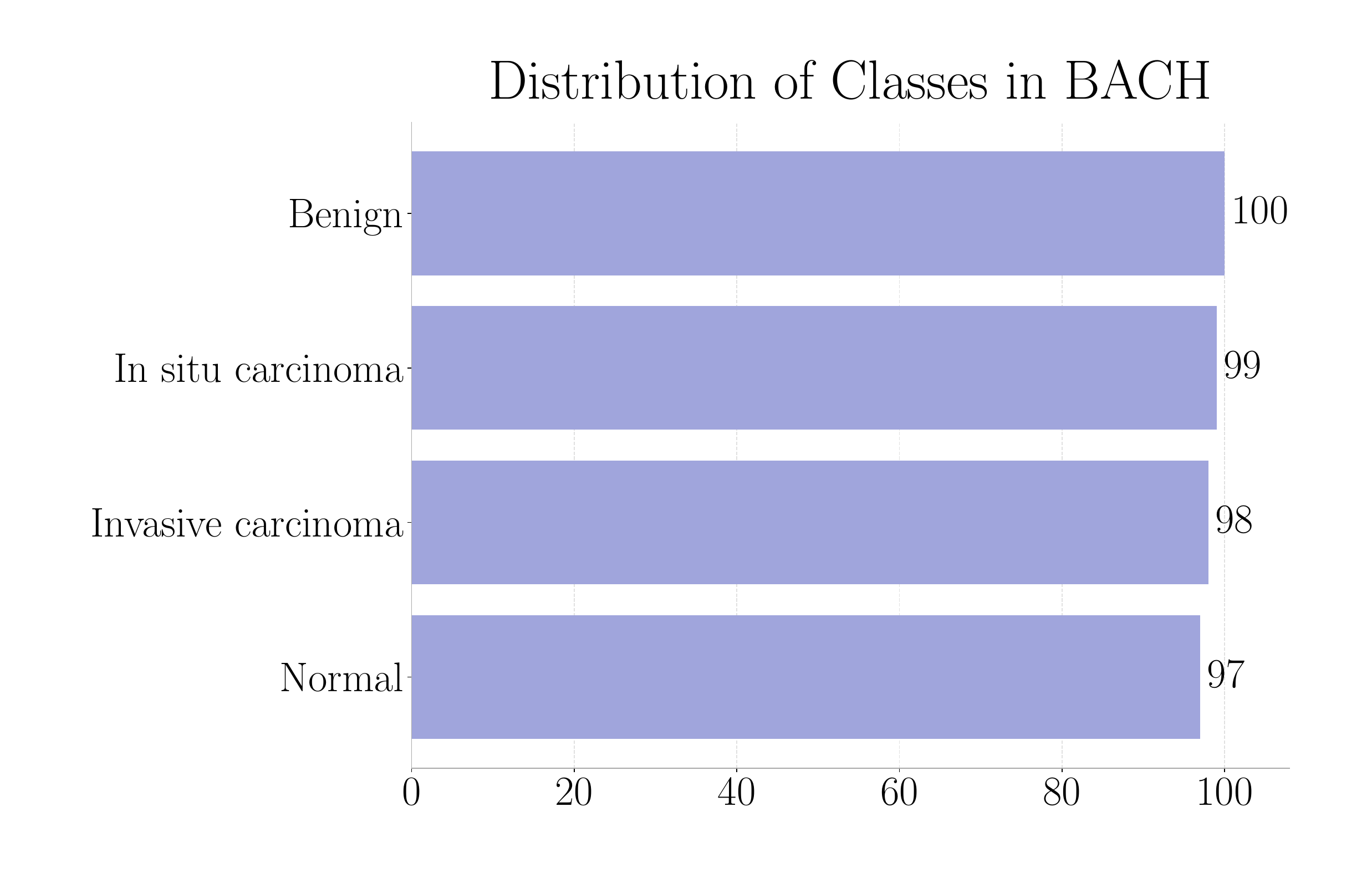}
    \end{subfigure}
    \begin{subfigure}{0.49\textwidth}
        \includegraphics[width=\linewidth]{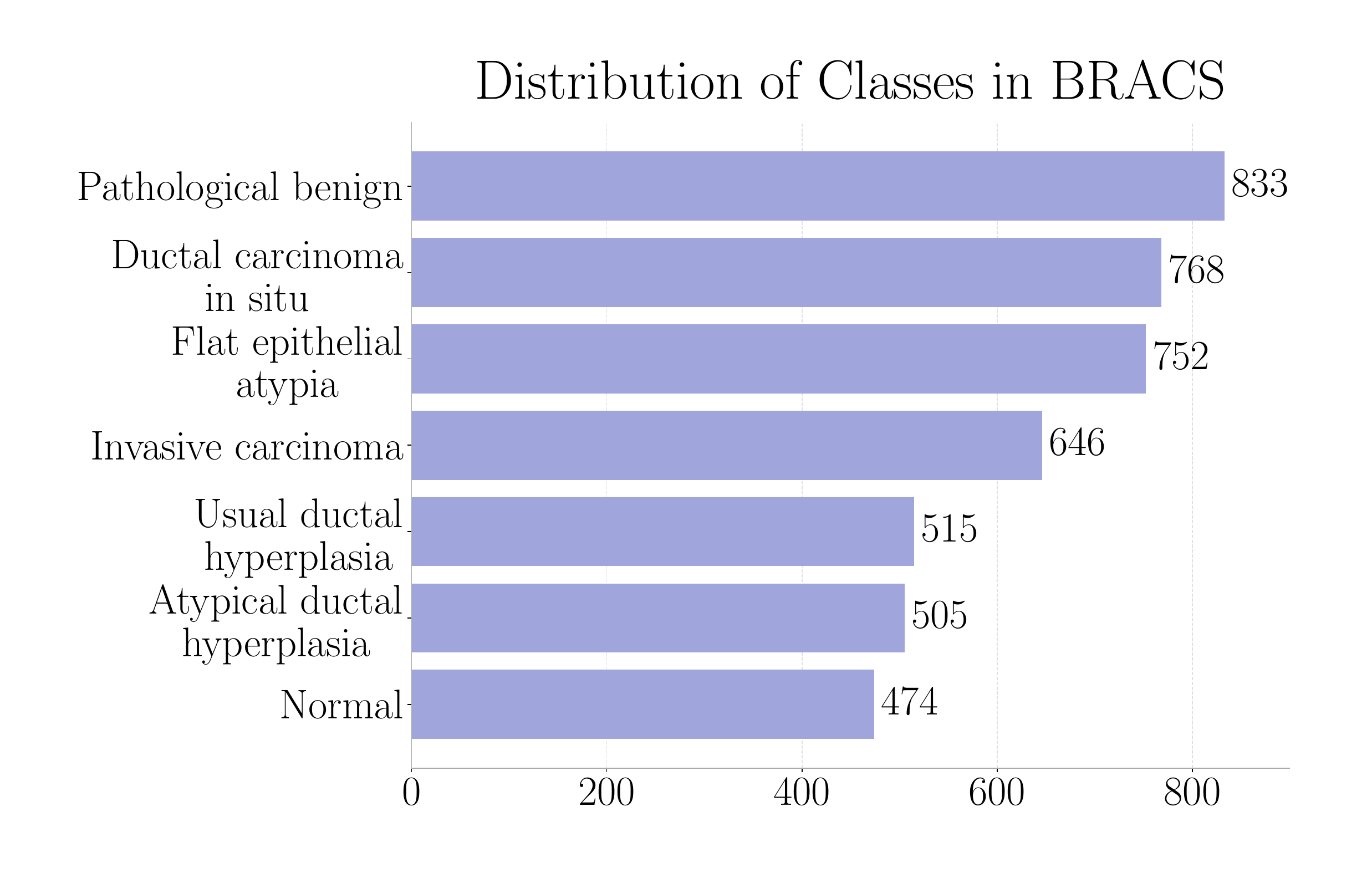}
    \end{subfigure}
    \begin{subfigure}{0.49\textwidth}
        \includegraphics[width=\linewidth]{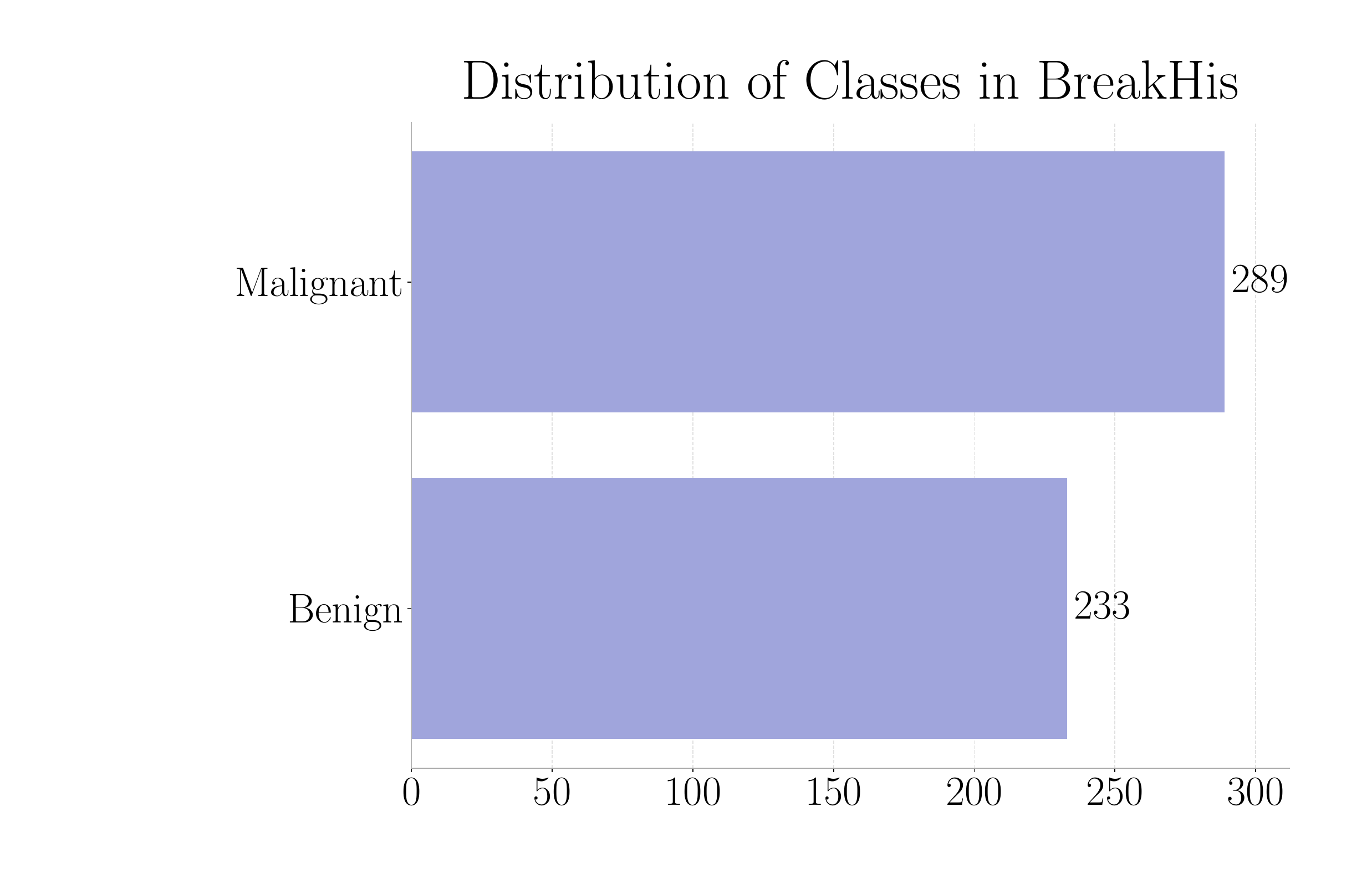}
    \end{subfigure}
    \begin{subfigure}{0.49\textwidth}
        \includegraphics[width=\linewidth]{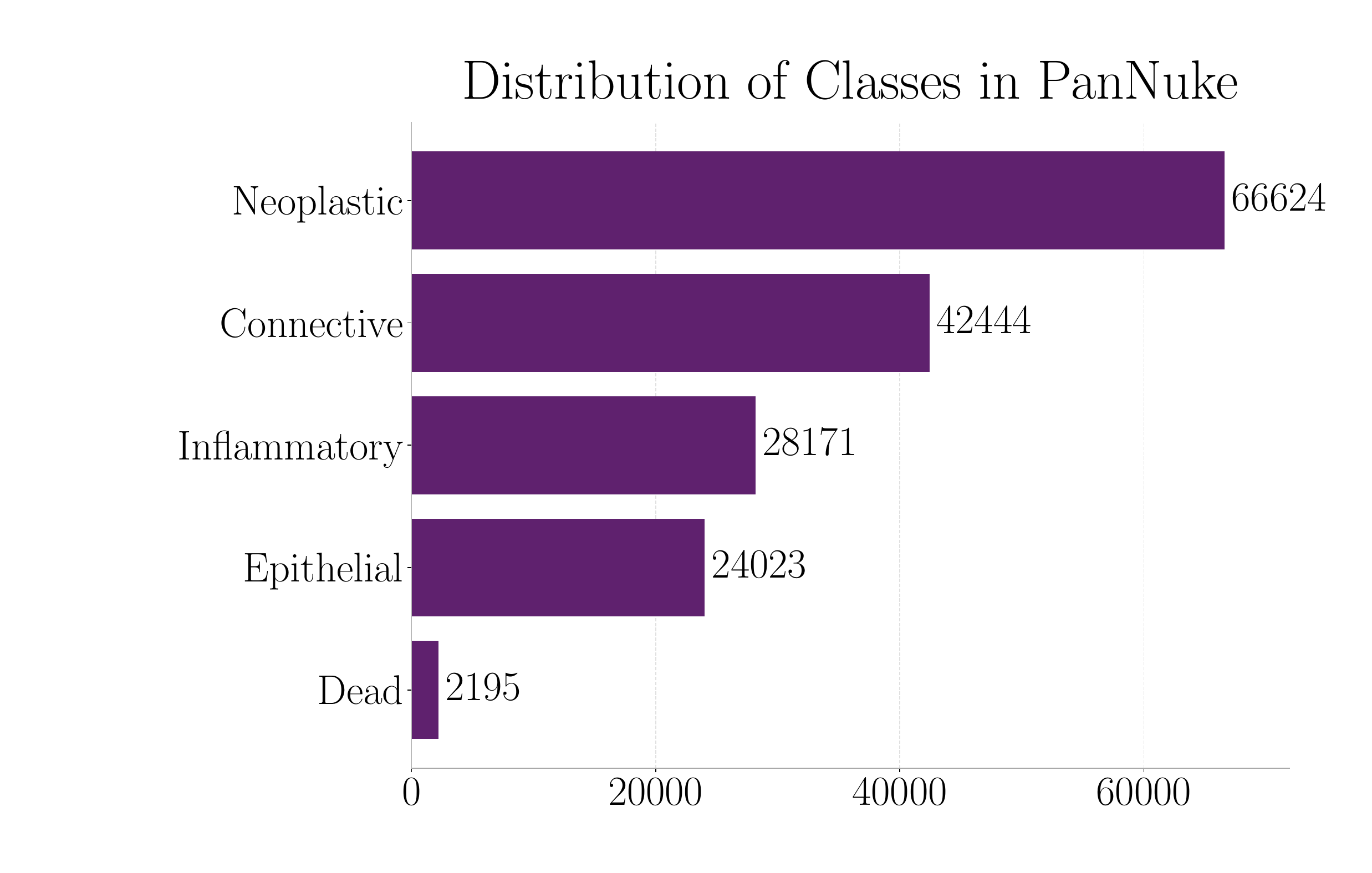}
    \end{subfigure}
    \begin{subfigure}{0.49\textwidth}
        \includegraphics[width=\linewidth]{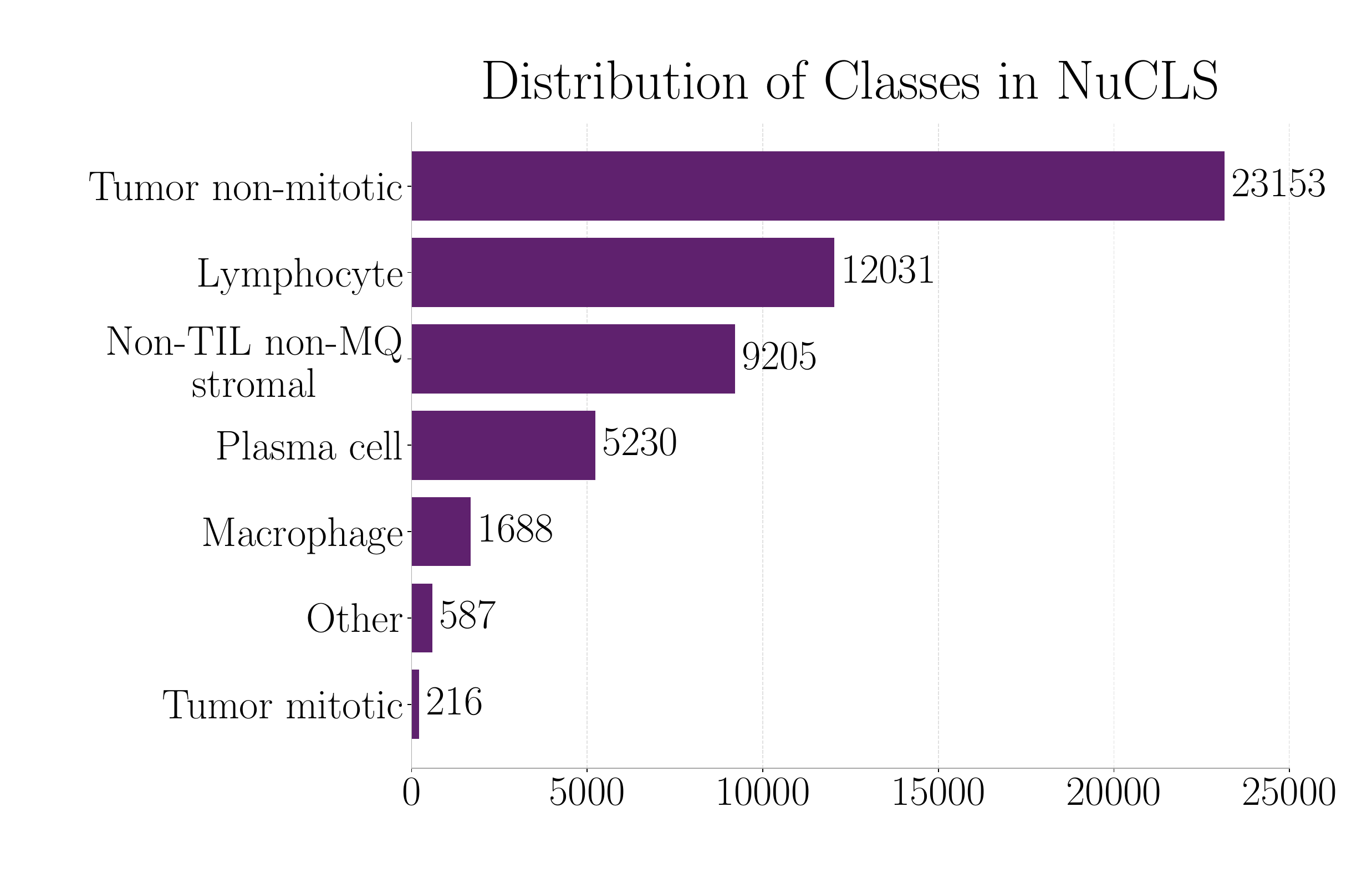}
    \end{subfigure}
    \caption{Class label distributions of \textcolor{slide}{slide-}, \textcolor{region}{region-}, and \textcolor{cell}{cell-}level datasets.}
    \label{fig:label_dist}
\end{figure}

Here, we describe the original image datasets.
To process these, we first employ the approach of~\citet{campanella2019clinical} to segment tissue regions from their background.
Then, unless otherwise specified, we adopt standard pre-processing conventions used in digital pathology foundation models, which consist of rescaling the segmented images at 20$\times$ magnification (i.e., 0.5 $\mu$m/pixel) before patching them into non-overlapping 224$\times$224-pixel tiles.
The statistics (Table~\ref{tab:dataset_stats}) and label distributions (Figure~\ref{fig:label_dist}) for each dataset are reported.

\begin{table}[t]
\caption{Statistics for the graph-based slide-, region-, and cell-level datasets.}
\label{tab:dataset_stats}
\centering
\resizebox{\textwidth}{!}{%
\begin{tabular}{@{}lrrrrrl@{}}
    \toprule
     & \textbf{\# Slides} & \textbf{\# Patches} & \textbf{Avg \# Nodes} & \textbf{Avg \# Edges} & \textbf{\# Classes} & \textbf{Tasks} \\
    \midrule
    \multicolumn{7}{@{}l}{\textbf{Slide-level}} \\
    TCGA-BRCA~\citep{weinstein2013cancer} & 998 & 11\,149\,500 & 43.94 & 119.16 & 2 & \shortstack[l]{Tumor subtyping\\Survival analysis} \\
    \midrule
    \multicolumn{7}{@{}l}{\textbf{Region-level}} \\
    BACH~\citep{aresta2019bach} & 394 & 14\,341 & 35.05 & 92.93 & 4 & \multirow{3}{*}{Tumor subtyping} \\
    BRACS~\citep{brancati2022bracs} & 4493 & 96\,153 & 46.96 & 128.09 & 7 & \\
    BreakHis~\citep{spanhol2015dataset} & 522 & 709 & 17.26 & 41.43 & 2 & \\
    \midrule
    \multicolumn{7}{@{}l}{\textbf{Cell-level}} \\
    PanNuke 40$\times$ (Breast)~\citep{gamper2020pannukedatasetextensioninsights} & --- & 2\,259 & 20.46 & 50.81 & 5 & \\
    PanNuke 40$\times$ (PanCancer)~\citep{gamper2020pannukedatasetextensioninsights} & --- & 7\,208 & 22.64 & 57.13 & 5 & \\
    NuCLS (main/super)~\citep{Amgad_2022} & --- & 1694 & 30.76 & 78.62 & 7/4 & \\
    \bottomrule
\end{tabular}}
\end{table}

\subsection{Slide-level}

\paragraph{TCGA-BRCA.}
We leverage this breast invasive carcinoma dataset, which comprises 998 H\&E-stained WSIs, sized up to 100K$\times$100K pixels~\citep{weinstein2013cancer}.
We obtain approximately 11 million patches from this dataset that serve two purposes in our study: (1) self-supervised pre-training to learn vision and graph representations, and (2) in-domain downstream evaluation corresponding to the classification of infiltrating ductal carcinoma and lobular carcinoma of different grades.
Note that multiple other subtypes were excluded in this study due to their small sample size.

\subsection{Region-level}

\paragraph{BACH.}
We use this breast cancer histology dataset~\citep{aresta2019bach} of 394 region-of-interest (RoI) images for out-of-domain downstream evaluation.
The pre-trained model embeddings are used for a four-class classification task of cancer subtyping distributed as normal, benign, in situ carcinoma, and invasive carcinoma.

\paragraph{BRACS.}
Similarly, we use this breast cancer dataset put forth by~\citet{brancati2022bracs} of 4493 labeled RoIs for downstream evaluation.
The end task consists of a seven-class classification of tumor subtypes, distributed as normal, pathological benign, usual ductal hyperplasia, flat epithelial atypia, atypical ductal hyperplasia, ductal carcinoma in situ, and invasive carcinoma.

\paragraph{BreakHis.}
The final region-level out-of-domain setting contains a dataset of 522 labeled microscopic images of breast tumor tissue collected from 82 patients~\citep{spanhol2015dataset}.
The downstream task is a binary classification of tumor type, with the labels of benign and malignant.

\subsection{Cell-level}

\paragraph{PanNuke.}
This cell-level dataset contains 200K annotated nuclei from 7904 patches and spans 19 distinct tissue types~\citep{gamper2020pannukedatasetextensioninsights}.
The end task provides granular classifications for nuclei into five categories of neoplastic, inflammatory, connective tissue, dead, and epithelial.
We use the predefined three folds.

\paragraph{NuCLS.}
Finally, this dataset consists of over 220K annotated nuclei designed for three classification tasks~\citep{Amgad_2022}.
The labels are organized into raw, main, and super sets, which represent varying levels of detail.
The raw one features 13 fine-grained classes, while the main and super sets provide categorizations of 7 and 4 classes, respectively, by aggregating the more detailed labels into broader groups.
We use only the main and super sets in this study.

\section{Pre-processing details}
\label{sec:pre_process_details}

We use the features shown in Table~\ref{tab:node_features} as node attributes and those shown in Table~\ref{tab:edge_features} as edge attributes for our cell graphs.
Please refer to~\citet{zhang2001comparative} for a detailed explanation of Fourier features with centroid signature and to~\citet{vadori2025automated} for their use as cell shape descriptors.
Additionally, Table~\ref{tab:preproc_runtime} reports the average pre-processing runtime.

\begin{table}[ht!]
\caption{The set of node-level color, morphology, and texture descriptors used in our graph construction pipeline. R, G, and B stand for red, green, and blue, respectively.}
\label{tab:node_features}
\tiny
\vskip 0.1in
\centering
\begin{tabular}{lll}
\toprule
\textbf{Nuclei-Intensity} & \textbf{Nuclei-Morph} & \textbf{Nuclei-Texture} \\
\midrule
min intensity R            & probability              & mean ASM \\
min intensity G            & orientation              & mean contrast \\
min intensity B            & axis major length        & mean correlation \\
max intensity R            & axis minor length        & mean dissimilarity \\
max intensity G            & eccentricity             & mean energy \\
max intensity B            & area                     & mean homogeneity \\
mean intensity R           & perimeter                & std ASM \\
mean intensity G           & circularity              & std contrast \\
mean intensity B           & elongation               & std correlation \\
std intensity R            & solidity                 & std dissimilarity \\
std intensity G            & extent                   & std energy \\
std intensity B            & Fourier descriptor 20    & std homogeneity \\
skew intensity R           & Fourier descriptor 30    & skew ASM \\
skew intensity G           &                          & skew contrast \\
skew intensity B           &                          & skew correlation \\
kurtosis intensity R       &                          & skew dissimilarity \\
kurtosis intensity G       &                          & skew energy \\
kurtosis intensity B       &                          & skew homogeneity \\
mean intensity gray scale  &                          & kurtosis ASM \\
std intensity gray scale   &                          & kurtosis contrast \\
skew intensity gray scale  &                          & kurtosis correlation \\
kurtosis intensity gray scale &                       & kurtosis dissimilarity \\
min intensity gray scale   &                          & kurtosis energy \\
max intensity gray scale   &                          & kurtosis homogeneity \\
                           &                          & min ASM \\
                           &                          & min contrast \\
                           &                          & min correlation \\
                           &                          & min dissimilarity \\
                           &                          & min energy \\
                           &                          & min homogeneity \\
                           &                          & max ASM \\
                           &                          & max contrast \\
                           &                          & max correlation \\
                           &                          & max dissimilarity \\
                           &                          & max energy \\
                           &                          & max homogeneity \\
\bottomrule
\end{tabular}
\end{table}
\begin{table}[ht!]
\caption{The set of edge-level color, morphology, texture, and relational descriptors used in our graph construction pipeline. R, G, and B stand for red, green, and blue, respectively, and ECM stands for extracellular matrix.}
\label{tab:edge_features}
\tiny
\vskip 0.1in
\centering
\begin{tabular}{llll}
\toprule
\textbf{ECM-Intensity} &
\textbf{ECM-Morph} &
\textbf{ECM-Texture} &
\textbf{ECM-Spatial/Relational} \\
\midrule
min intensity R              & axis major length & mean ASM             & distance ($\mu$m) \\
min intensity G              & axis minor length & mean contrast        & $\cos(2\theta)$ \\
min intensity B              & area              & mean correlation     & $\sin(2\theta)$ \\
max intensity R              & perimeter         & mean dissimilarity   & cell size ratio \\
max intensity G              & circularity       & mean energy          & color contrast R \\
max intensity B              & elongation        & mean homogeneity     & color contrast G \\
mean intensity R             &                   & std ASM              & color contrast B \\
mean intensity G             &                   & std contrast         & texture contrast contrast \\
mean intensity B             &                   & std correlation      & texture contrast homogeneity \\
std intensity R              &                   & std dissimilarity    & \\
std intensity G              &                   & std energy           & \\
std intensity B              &                   & std homogeneity      & \\
skew intensity R             &                   & skew ASM             & \\
skew intensity G             &                   & skew contrast        & \\
skew intensity B             &                   & skew correlation     & \\
kurtosis intensity R         &                   & skew dissimilarity   & \\
kurtosis intensity G         &                   & skew energy          & \\
kurtosis intensity B         &                   & skew homogeneity     & \\
mean intensity gray scale    &                   & kurtosis ASM         & \\
std intensity gray scale     &                   & kurtosis contrast    & \\
skew intensity gray scale    &                   & kurtosis correlation & \\
kurtosis intensity gray scale&                   & kurtosis dissimilarity & \\
min intensity gray scale     &                   & kurtosis energy      & \\
max intensity gray scale     &                   & kurtosis homogeneity & \\
                             &                   & min ASM              & \\
                             &                   & min contrast         & \\
                             &                   & min correlation      & \\
                             &                   & min dissimilarity    & \\
                             &                   & min energy           & \\
                             &                   & min homogeneity      & \\
                             &                   & max ASM              & \\
                             &                   & max contrast         & \\
                             &                   & max correlation      & \\
                             &                   & max dissimilarity    & \\
                             &                   & max energy           & \\
                             &                   & max homogeneity      & \\
\bottomrule
\end{tabular}
\end{table}

\section{Pre-training details}
\label{sec:pre_train_details}

\begin{figure}[htbp]
\centering
\setlength{\abovecaptionskip}{6pt}
\begin{minipage}[t]{0.46\textwidth}
\vspace{0pt}
\begin{minipage}[c][4.5cm][c]{\linewidth}
\centering
\small
\begin{tabular*}{\linewidth}{@{\extracolsep{\fill}}lc@{}}
\toprule
\textbf{Model} & \shortstack{\textbf{Number of Parameters}\\($10^6$)} \\
\midrule
DINOv2               & 22.01 \\
MAE                  & 47.58 \\
\midrule
GrapHist-S ($d=512$)  & \textbf{9.50} \\
GrapHist-S ($d=768$)  & 21.12 \\
GrapHist-S ($d=1024$) & 37.34 \\
GrapHist-E ($d=512$)  & 10.53 \\
\bottomrule
\end{tabular*}
\end{minipage}
\captionof{table}{Model size comparison across SSL methods. Here $d$ denotes the validated embedding dimension of GrapHist-S and GrapHist-E.}
\label{tab:num_params}
\end{minipage}\hfill
\begin{minipage}[t]{0.50\textwidth}
\vspace{0pt}
\begin{minipage}[c][4.5cm][c]{\linewidth}
\centering
\includegraphics[width=\linewidth,height=4.5cm,keepaspectratio]{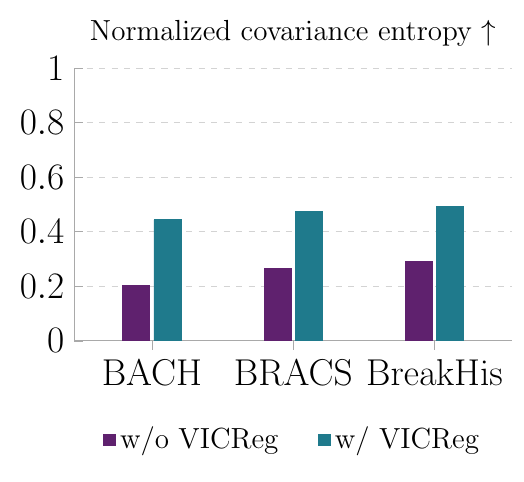}
\end{minipage}
\captionof{figure}{Normalized entropy of the full PCA explained variance spectrum for GrapHist-E trained with and without VICReg.}
\label{fig:vicreg_collapse_metrics}
\end{minipage}
\end{figure}

We pre-train GrapHist-S on the TCGA-BRCA cell graphs for $100$ epochs with Adam, using a learning rate of $10^{-3}$, a batch size of $2048$, no weight decay, no scheduler, and seed $0$.
The encoder comprises five layers and the decoder one, both instantiated with the architecture of Section~\ref{sec:backbone} with $d = 512$, PReLU activations, and no normalization layers.
Masking is applied at a rate $r_m = 0.5$ with a replacement rate $r_r = 0.1$, and the scaled cosine error uses $\gamma = 3$.
GrapHist-E inherits this configuration but differs from it in the following two aspects.
First, each edge carries a $75$-dimensional attribute whose first entry is the centroid-to-centroid distance.
Second, to prevent dimensional collapse, we augment the loss with the variance and covariance terms of VICReg~\citep{bardes2022vicreg}, weighted $0.05$ and $0.002$, respectively.
To motivate the use of the VICReg loss, we measure the normalized Shannon entropy of the full PCA explained variance spectrum of the centered patch embeddings.
As shown in Figure~\ref{fig:vicreg_collapse_metrics}, adding VICReg increases this entropy across BACH, BRACS, and BreakHis.
This indicates a more even distribution of variance across embedding directions, consistent with reduced dimensional collapse.
Finally, all experiments were run on NVIDIA H200 (140GB) GPUs, and the number of parameters in DINOv2, MAE, GrapHist-S, and GrapHist-E models can be found in Table~\ref{tab:num_params}.
We provide the specific setup used in the evaluation of our self-supervised and fully-supervised experiments in the paragraphs below, along with the validated hyperparameters.
\newline

\section{Inference details}
\label{sec:inference_details}

\subsection{Setup for SSL evaluation}
\label{sec:setup_ssl}

We remark that for slide- and region-level tasks, we employ a five-fold stratified cross-validation on the training set to find the best hyperparameters of the MIL heads.
In detail, we conduct a grid search over the learning rate in $\{0.001, 0.01\}$, dropout in $\{0.2, 0.5\}$, hidden dimension of the attention in $\{128, 256\}$, and the number of layers in the classifier in $\{1, 2\}$.
The hyperparameter configuration yielding the highest mean validation macro F1 score across the folds is selected, and the corresponding models are then evaluated on the held-out test set.
We repeat this procedure for three MIL aggregators (ABMIL, add-ABMIL, and conj-ABMIL) and report the results of the best-performing variant in Table~\ref{tab:tumor_subtyping}.
A detailed breakdown of the performance for each individual aggregator is provided in the next section.
For cell-level tasks, we use the provided train/test splits of PanNuke and NuCLS datasets and use logistic regression for linear probing.

\subsection{Setup for SL evaluation}
\label{sec:setup_sl}

For slide- and region-level tasks, we employ an ACM model with an MIL head.
We select the optimal hyperparameters for the GNN's hidden dimension in $\{16, 32\}$, number of layers in $\{1, 2\}$, and hidden dimension of the attention in $\{64, 128\}$ using a stratified five-fold cross-validation based on the highest mean validation F1 score.
For cell-level tasks, we similarly use an ACM backbone with an MLP classifier on the node level and validate the GNN's hidden dimension in $\{256, 512\}$, its number of layers in $\{3, 5\}$, and the MLP's hidden dimension in $\{64, 128\}$ using a two-fold and four-fold cross-validation for PanNuke and NuCLS, respectively, given the original splits in each dataset.

\section{Fine-tuning details}
\label{sec:fine_tune_details}

Below, we provide the details of the parameter-efficient fine-tuning implemented in Section~\ref{sec:cell_phenotyping}. 

\subsection{Adapter architecture}

We adapt the pre-trained GrapHist-S encoder with a two-branch bottleneck adapter following AdapterGNN~\citep{li2024adaptergnn}, but assign its branch weights from the channel-mixing coefficients.
Collecting the terms of Eq.~\ref{eq:acm} by signal reveals that, up to the per-channel MLPs, an ACM layer combines only two quantities for a given cell $v_i$, which are its own features $\vh_i$ and its aggregated neighborhood $\vm_i$.
Writing $\alpha_{i,L}$, $\alpha_{i,H}$, and $\alpha_{i,I}$ for the mixing coefficients that Eq.~\ref{eq:acm} assigns to $v_i$, the weights of these two quantities are $b_1 = 1 - \tfrac{\alpha_{i,L} + \alpha_{i,H}}{2}$ and $b_2 = \tfrac{\alpha_{i,L} - \alpha_{i,H}}{2}$.
Concretely, we add to the output of layer $k$ the correction $b_1 \, g_1(\vh_i^{(k-1)}) + b_2 \, g_2(\vm_i^{(k-1)})$, where each $g$ is a bottleneck block composed of a down-projection, a ReLU, a zero-initialized up-projection, and batch normalization.
The coefficients are detached, so that they guide the adapter without being trained through it.

\subsection{Setup for FT evaluation}

We follow the supervised cell-level protocol of Appendix~\ref{sec:inference_details} with the pre-trained checkpoint held fixed.
The bottleneck dimension $\in \{16, 64\}$, the MLP hidden dimension $\in \{64, 128\}$, and the classifier dropout $\in \{0.3, 0.5\}$ are selected by mean validation macro F1.
We train for $250$ epochs with Adam at a learning rate of $10^{-5}$, halved every $25$ epochs, and seed $0$.

\section{Additional results}

Here, we provide results from individual MIL aggregators for slide- and region-level tasks (Table~\ref{tab:detailed_mil}), analyze the homophily/heterophily properties of our cell-level datasets, and conduct a sensitivity analysis for GrapHist-S to critical hyperparameters.

\subsection{Homophily/heterophily analysis}

\begin{figure}[htbp]
\centering
\setlength{\abovecaptionskip}{6pt}
\begin{minipage}[t]{0.46\textwidth}
\vspace{0pt}
\begin{minipage}[c][4.5cm][c]{\linewidth}
\centering
\small
\begin{tabular*}{\linewidth}{@{\extracolsep{\fill}}lcc@{}}
\toprule
\textbf{Dataset} & \textbf{$H_v<0.3$ (\%)} & \textbf{$H_v<0.5$ (\%)} \\
\midrule
PanNuke 20$\times$ & 10.46 & 16.83 \\
PanNuke 40$\times$ & 10.48 & 16.83 \\
NuCLS-main         & 7.93  & 14.32 \\
NuCLS-super        & 5.55  & 10.77 \\
\bottomrule
\end{tabular*}
\end{minipage}
\captionof{table}{Per-node homophily $H_v$ across the downstream cell-level datasets.}
\label{tab:homophily}
\end{minipage}\hfill
\begin{minipage}[t]{0.50\textwidth}
\vspace{0pt}
\begin{minipage}[c][4.5cm][c]{\linewidth}
\centering
\includegraphics[width=\linewidth,height=4.5cm,keepaspectratio]{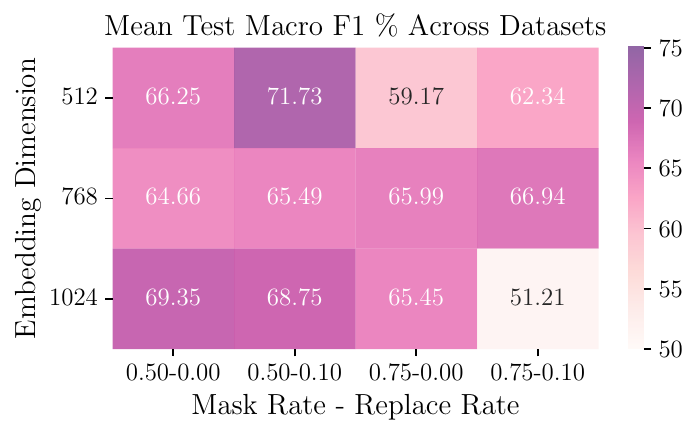}
\end{minipage}
\captionof{figure}{Mean macro F1 across the OOD region-level tasks for different pre-training configurations.}
\label{fig:sensitivity_analysis}
\end{minipage}
\end{figure}

We investigate whether the graphs used in our framework actually exhibit the properties that motivate our backbone, which combines homophilic and heterophilic components.
As such, we focus on our cell-level datasets, where node labels are already available, and enable us to quantify the neighborhood homophily/heterophily for a given graph.
Specifically, we adhere to the metric introduced by~\citet{luan2022revisiting}, defined as $H_v = |\{u \in \mathcal{N}_v : y_u = y_v\}| / d_v$, where $\mathcal{N}_v$ is the set of neighbors of node $v$, $d_v$ is its degree, and $y_v$ is its label.
Simply put, $H_v$ quantifies the fraction of a cell's neighbors that share the same label as itself.
For each cell-level dataset, we report the fraction of cells below the standard heterophily cutoff of $H_v<0.5$~\citep{pei2020geom} and a stricter $H_v<0.3$.
We observe in Table~\ref{tab:homophily} that on the labelled datasets, $5.6$--$10.5\%$ of cells have $H_v<0.3$ and $10.8$--$16.8\%$ have $H_v<0.5$, which is primarily due to rare cell types.
This is indeed the regime that our channel mixing targets.
The low-pass channel serves the homophilic majority, while the high-pass one preserves the distinguishing signal of the heterophilic minority.
\newline

\subsection{Sensitivity analysis}
\label{sec:sensitivity}

Finally, we study the sensitivity of GrapHist-S to its validated hyperparameters on the out-of-distribution region-level benchmarks.
For each configuration, we report the test macro F1 performance achieved on average across MIL variants in Figure~\ref{fig:sensitivity_analysis}.
We can observe a notable performance gap of 8-20\% between the tested configurations.
Nonetheless, the overall hyperparameter sensitivity dynamics are consistent with findings of~\citet{hou2022graphmae}, even though we operate at a significantly larger scale than their setting.
Our best-performing model coincides with the lowest masking rate and the highest replacement rate as seen in Figure~\ref{fig:sensitivity_analysis}.
The best embedding dimension matches several molecular datasets whose graphs have around a hundred continuous node features.
\newline

\begin{table}[htbp]
\caption{Test macro F1 scores (\%) on slide- and region-level tumor subtyping for all evaluated models and aggregation methods, reported as mean $\pm$ standard deviation over the five cross-validation models evaluated on the held-out test set. Best and second-best results are highlighted in bold and underlined, respectively.}
\centering
\resizebox{\textwidth}{!}{%
\begin{tabular}{@{}ll@{\hspace{10pt}}c@{\hspace{10pt}}c@{\hspace{10pt}}c@{\hspace{10pt}}c@{}}
\toprule
& & \multicolumn{4}{c}{\textbf{Datasets}} \\
\cmidrule(lr){3-6}
\textbf{Model} & \textbf{Aggregation} & \multicolumn{1}{c}{Slide} & \multicolumn{3}{c}{Region} \\
\cmidrule(lr){3-3}\cmidrule(lr){4-6}
 &  & TCGA-BRCA & BACH & BRACS & BreakHis \\
\midrule
\textbf{Supervised graph models} & & & & & \\
\multirow{3}{*}{ACM-bio} & ABMIL & — & 19.56 \scriptsize{$\pm$ 5.32} & 11.80 \scriptsize{$\pm$ 1.76} & 55.82 \scriptsize{$\pm$ 10.33} \\
 & add-ABMIL & — & 38.37 \scriptsize{$\pm$ 2.16} & 21.55 \scriptsize{$\pm$ 2.62} & 49.30 \scriptsize{$\pm$ 2.53} \\
 & conj-ABMIL & — & 19.63 \scriptsize{$\pm$ 3.89} & 13.89 \scriptsize{$\pm$ 4.23} & 53.61 \scriptsize{$\pm$ 10.41} \\
\midrule
\multirow{3}{*}{ACM-UNI} & ABMIL & — & 22.26 \scriptsize{$\pm$ 9.08} & 9.62 \scriptsize{$\pm$ 0.76} & 70.72 \scriptsize{$\pm$ 6.50} \\
 & add-ABMIL & — & 29.03 \scriptsize{$\pm$ 9.92} & 19.95 \scriptsize{$\pm$ 3.27} & 69.22 \scriptsize{$\pm$ 8.22} \\
 & conj-ABMIL & — & 24.52 \scriptsize{$\pm$ 10.40} & 12.17 \scriptsize{$\pm$ 3.57} & 76.27 \scriptsize{$\pm$ 9.23} \\
\midrule
\textbf{Self-supervised vision models} & & & & & \\
\multirow{3}{*}{DINOv2} & ABMIL & 62.43 \scriptsize{$\pm$ 1.75} & 55.15 \scriptsize{$\pm$ 3.28} & 47.88 \scriptsize{$\pm$ 1.63} & 81.06 \scriptsize{$\pm$ 4.40} \\
 & add-ABMIL & 58.41 \scriptsize{$\pm$ 4.32} & 59.25 \scriptsize{$\pm$ 3.34} & 52.68 \scriptsize{$\pm$ 1.78} & 83.44 \scriptsize{$\pm$ 2.25} \\
 & conj-ABMIL & 62.21 \scriptsize{$\pm$ 2.54} & 48.95 \scriptsize{$\pm$ 1.19} & 46.25 \scriptsize{$\pm$ 2.11} & 82.21 \scriptsize{$\pm$ 4.55} \\
\midrule
\multirow{3}{*}{MAE} & ABMIL & 66.72 \scriptsize{$\pm$ 1.27} & 53.30 \scriptsize{$\pm$ 2.24} & 51.00 \scriptsize{$\pm$ 0.93} & 87.02 \scriptsize{$\pm$ 2.51} \\
 & add-ABMIL & 60.65 \scriptsize{$\pm$ 3.10} & 57.94 \scriptsize{$\pm$ 3.91} & 56.33 \scriptsize{$\pm$ 0.76} & \underline{87.74 \scriptsize{$\pm$ 1.66}} \\
 & conj-ABMIL & 66.35 \scriptsize{$\pm$ 2.83} & 57.25 \scriptsize{$\pm$ 3.20} & 50.91 \scriptsize{$\pm$ 0.92} & 87.52 \scriptsize{$\pm$ 2.19} \\
\midrule
\textbf{Ours} & & & & & \\
\multirow{3}{*}{GrapHist-S} & ABMIL & 71.40 \scriptsize{$\pm$ 2.73} & \textbf{69.16 \scriptsize{$\pm$ 3.37}} & 58.70 \scriptsize{$\pm$ 0.97} & 85.84 \scriptsize{$\pm$ 3.82} \\
 & add-ABMIL & 70.17 \scriptsize{$\pm$ 2.55} & 68.77 \scriptsize{$\pm$ 4.60} & 60.30 \scriptsize{$\pm$ 0.46} & 87.27 \scriptsize{$\pm$ 4.40} \\
 & conj-ABMIL & \textbf{72.25 \scriptsize{$\pm$ 1.40}} & 66.74 \scriptsize{$\pm$ 2.83} & 59.39 \scriptsize{$\pm$ 1.89} & 89.37 \scriptsize{$\pm$ 1.94} \\
 \midrule
 \multirow{3}{*}{GrapHist-E} & ABMIL & \underline{72.06 \scriptsize{$\pm$ 2.50}} & 64.20 \scriptsize{$\pm$ 3.15} & 65.80 \scriptsize{$\pm$ 1.72} & \textbf{95.53 \scriptsize{$\pm$ 2.44}} \\
 & add-ABMIL & 65.42 \scriptsize{$\pm$ 4.38} & \underline{68.98 \scriptsize{$\pm$ 3.59}} & \textbf{69.00 \scriptsize{$\pm$ 1.30}} & \underline{94.35 \scriptsize{$\pm$ 2.12}} \\
 & conj-ABMIL & 71.97 \scriptsize{$\pm$ 2.30} & 65.57 \scriptsize{$\pm$ 2.89} & \underline{67.91 \scriptsize{$\pm$ 1.85}} & 94.18 \scriptsize{$\pm$ 2.65} \\
\bottomrule
\end{tabular}
}
\label{tab:detailed_mil}
\end{table}

\end{document}